\let\TeXyear\year
\documentclass{ieeeaccess}
\let\setyear\year
\let\year\TeXyear

\usepackage{cite}
\usepackage{amsmath,amssymb,amsfonts}
\usepackage{algorithmic}
\usepackage{graphicx}
\usepackage{textcomp}

\usepackage{multirow}

\usepackage[nolist]{acronym}
\usepackage{hyperref}
\usepackage[binary-units=true]{siunitx}
\usepackage{tabularx}
\newcolumntype{Y}{>{\centering\arraybackslash}X}
\usepackage{booktabs}  % for \toprule, \midrule, and \bottomrule macros 
\usepackage{makecell}
\usepackage{lipsum}

\usepackage{color,soul}
\soulregister\acrlong7
\soulregister\ac7
\soulregister\acp7
\soulregister\ref7

\ExplSyntaxOn
\tl_new:N \l_hlautosplit_hlsplitstring_tl
\NewDocumentCommand{\hlautosplit} { m } {
    \tl_clear:N \l_hlautosplit_hlsplitstring_tl
    \str_map_inline:nn { #1 } {
        \tl_put_right:Nn \l_hlautosplit_hlsplitstring_tl { ##1 \soulomit{\hspace{0pt plus 0.1pt}} }
    }
    \exp_args:No \hl { \l_hlautosplit_hlsplitstring_tl }
}
\ExplSyntaxOff

\usepackage{mathtools}
\usepackage{bm}
\makeatletter
\AtBeginDocument{\DeclareMathVersion{bold}
\SetSymbolFont{operators}{bold}{T1}{times}{b}{n}
\SetSymbolFont{NewLetters}{bold}{T1}{times}{b}{it}
\SetMathAlphabet{\mathrm}{bold}{T1}{times}{b}{n}
\SetMathAlphabet{\mathit}{bold}{T1}{times}{b}{it}
\SetMathAlphabet{\mathbf}{bold}{T1}{times}{b}{n}
\SetMathAlphabet{\mathtt}{bold}{OT1}{pcr}{b}{n}
\SetSymbolFont{symbols}{bold}{OMS}{cmsy}{b}{n}
\renewcommand\boldmath{\@nomath\boldmath\mathversion{bold}}}
\makeatother

\def\BibTeX{{\rm B\kern-.05em{\sc i\kern-.025em b}\kern-.08em
    T\kern-.1667em\lower.7ex\hbox{E}\kern-.125emX}}

\usepackage{tikz}
  \NewSpotColorSpace{PANTONE}
  \AddSpotColor{PANTONE} {PANTONE3015C} {PANTONE\SpotSpace 3015\SpotSpace C} {1 0.3 0 0.2}
  \SetPageColorSpace{PANTONE}
 \usepackage{pgfplots}
 \pgfplotsset{compat=1.16}
 \usetikzlibrary{backgrounds,arrows,automata,shapes,positioning,calc,through}
 \pgfdeclarelayer{background}
 \pgfdeclarelayer{foreground}
 \pgfsetlayers{background,main,foreground}

 \usetikzlibrary{fit}

\tikzset{
  state/.style={
    rectangle,
    draw=black, very thick,
    minimum height=1.0em,
    text centered,
  },
  final_state/.style={
    rectangle,
    rounded corners,
    draw=black, very thick,
    minimum height=2em,
    text centered,
    dashed
  },
  initial_state/.style={
    rectangle,
    double=white,
    double distance=1pt,
    inner sep=2pt,
    draw=black, very thick,
    minimum height=2em,
    text centered,
  },
  point/.style={
    circle,
    inner sep=0pt,
    minimum size=3pt,
    fill=red
  }
}
 \usepackage{pstricks}
 \usepackage{tikz-3dplot}

\setyear{2026}

\renewcommand\vec{\bm}
\newcommand\mat{\mathbf}
\newcommand{\tstep}[2][]{_{#1[#2]}}

\newcommand{\frames}[3]{{\prescript{#1}{#2}{#3}}}
\usepackage{cancel}

\newcommand\rev{}
\newcommand\bl{}

\newcommand{\DOI}{https://doi.org/10.1109/ACCESS.2026.3666998} % you will not get a DOI until the paper is actually published, so update this when you get it and reupload the new preprint to all systems

\usepackage[placement=top,vshift=-7]{background}
\SetBgScale{1.0}
\SetBgContents{IEEE Access. Accepted version. \href{https://doi.org/\DOI}{DOI \DOI}}
\SetBgColor{black}
\SetBgAngle{0}
\SetBgOpacity{1.0}

\begin{document}

% % Place this in the document body as the first part to have an extra first page with the copyright. Remove this part to get rid of the extra page.
% \thispagestyle{empty}
% \onecolumn
% {
%   \topskip0pt
%   \vspace*{\fill}
%   \centering
%   \LARGE{%
%     \copyright{} \PREPRINTYEAR~\PUBLISHEDIN\\\vspace{1cm}
%     Personal use of this material is permitted.
%     % Permission from \PUBLISHEDIN~must be obtained for all other uses, in any current or future media, including reprinting or republishing this material for advertising or promotional purposes, creating new collective works, for resale or redistribution to servers or lists, or reuse of any copyrighted component of this work in other works.}
%     Permission from IEEE~must be obtained for all other uses, in any current or future media, including reprinting or republishing this material for advertising or promotional purposes, creating new collective works, for resale or redistribution to servers or lists, or reuse of any copyrighted component of this work in other works.}
%     \vspace*{\fill}
% }
% \NoBgThispage
% \twocolumn          	% Comment out for single-column articles
% \BgThispage

\history{Date of publication xxxx 00, 0000, date of current version xxxx 00, 0000.}
\doi{10.1109/ACCESS.2026.3666998}

\title{Fusion of Visual-Inertial Odometry with LiDAR Relative Localization for Cooperative Guidance of a Micro-Scale Aerial Vehicle}
\author{\uppercase{V\'aclav Pritzl}\authorrefmark{1},
\uppercase{Matou\v{s} Vrba}\authorrefmark{1}, \uppercase{Petr \v{S}t\v{e}p\'an}\authorrefmark{1}, and \uppercase{Martin Saska}\authorrefmark{1},
\IEEEmembership{Member, IEEE}}

\address[1]{Multi-robot Systems Group, Department of Cybernetics, Faculty of Electrical Engineering, Czech Technical University in Prague, Technick\'a 2, Prague, 166 27, Czech Republic}
\tfootnote{This work was funded by the Czech Science Foundation (GA\v{C}R) under research project no. 24-12360S, and by the European Union under the project Robotics and advanced industrial production (reg. no. CZ.02.01.01/00/22\_008/0004590).}

\markboth
{V. Pritzl \headeretal: Fusion of VIO with LiDAR Relative Localization for Cooperative Guidance of a Micro-Scale Aerial Vehicle}
{V. Pritzl \headeretal: Fusion of VIO with LiDAR Relative Localization for Cooperative Guidance of a Micro-Scale Aerial Vehicle}

\corresp{Corresponding author: V\'aclav Pritzl (e-mail: vaclav.pritzl@fel.cvut.cz).}

\begin{abstract}
  A novel relative localization approach for guidance of a micro-scale \ac{UAV} by a well-equipped aerial robot fusing \ac{VIO} with \ac{lidar} is proposed in this paper.
  \ac{lidar}-based localization is accurate and robust to challenging environmental conditions, but 3D \acp{lidar} are relatively heavy and require large \ac{UAV} platforms, in contrast to lightweight cameras.
  However, visual-based self-localization methods exhibit lower accuracy and can suffer from significant drift with respect to the global reference frame.
  To benefit from both sensory modalities, we focus on cooperative navigation in a heterogeneous team of a primary \ac{lidar}-equipped \ac{UAV} and a secondary micro-scale camera-equipped \ac{UAV}.
  We propose a novel cooperative approach combining \ac{lidar} relative localization data with \ac{VIO} output on board the primary \ac{UAV} to obtain an accurate pose of the secondary \ac{UAV}.
  The pose estimate is used to precisely and reliably guide the secondary \ac{UAV} along trajectories defined in the primary \ac{UAV} reference frame.
  The experimental evaluation has shown the superior accuracy of our method to the raw \ac{VIO} output\rev{, reaching the average 3D \ac{ATE} of {\SI{0.28}{m}}}, and demonstrated its capability to guide the secondary \ac{UAV} along desired trajectories while mitigating \ac{VIO} drift.
  Thus, such a heterogeneous system can explore large areas with \ac{lidar} precision, as well as visit locations inaccessible to the large \ac{lidar}-carrying \ac{UAV} platforms, as was showcased in a real-world cooperative mapping scenario.
\end{abstract}


\titlepgskip=-21pt

\maketitle

\section*{Multimedia Attachment}

\noindent\url{http://mrs.felk.cvut.cz/coop-fusion}

\section{Introduction}
\label{sec:introduction}
    % \vspace{-0.3em}

% INTRODUCTION%%{
\acresetall

Autonomous multi-rotor \acp{UAV} are well-suited for emergency response and \ac{SAR} tasks in confined indoor spaces.
In these applications, the \acp{UAV} operate in \ac{GNSS}-denied environments but still require accurate positioning in a world reference frame, e.g., to precisely report victim positions, as was demonstrated in the \ac{DARPA} \ac{SubT} challenge~\cite{petrlikUAVsSurfaceCooperative2022, kratkyAutonomousUnmannedAerial2021}.

Camera-equipped \acp{UAV} can be localized by \ac{VIO} techniques and may be small enough to fly through highly confined environments.
However, visual-based localization approaches are heavily dependent on lighting conditions~\cite{BednarICUAS22VIO}, the amount of texture in the environment~\cite{limAvoidingDegeneracyMonocular2021}, and suffer from long-term drift with respect to the world frame due to the \ac{VIO}'s four unobservable \acp{DOF}~\cite{heschConsistencyAnalysisImprovement2014}.
Monocular camera-based methods exhibit scale unobservability under constant acceleration~\cite{wuUnobservableDirectionsVINS2016a, yangObservabilityAnalysisAided2019}.
The impracticality of purely vision-based approaches for indoor and subterranean \ac{SAR} missions was confirmed by all \ac{DARPA} \ac{SubT} teams, including the team of authors of this article.

\ac{lidar}-based methods rely on the geometric structure of the environment.
In comparison to visual-based methods, 3D \ac{lidar}-based localization exhibits lower drift and performs well under challenging lighting conditions.
Most of the best-performing algorithms on the \ac{KITTI} dataset~\cite{Geiger2012CVPR} are \ac{lidar}-based.
However, 3D \acp{lidar} are relatively heavy and power-consuming.
Although all the teams in the \ac{DARPA} SubT challenge relied on \acp{lidar}, the \ac{UAV} platforms needed for carrying such payloads were too bulky for exploring many narrow passages~\cite{ebadiPresentFutureSLAM2022}.

\begin{figure}
\centering
  \includegraphics[width=1.0\linewidth, trim=0.0cm 0.0cm 0.0cm 0.0cm, clip=true]{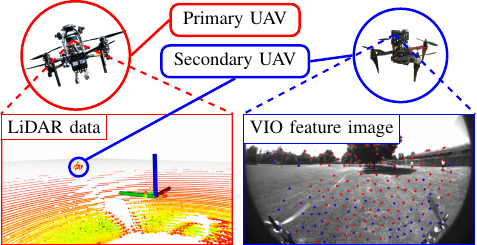}
  \caption{The primary \ac{UAV} carrying a 3D \ac{lidar} guides the secondary camera-equipped \ac{UAV} using a fusion of \ac{lidar} detections and \ac{VIO} data.}
\label{fig:motivation}
  % \vspace{-1.5em}
\end{figure}

Using teams of \acp{UAV} with different sensors keeps the benefits of the localization methods while mitigating the disadvantages.
The \ac{lidar}-equipped \ac{UAV} can share its localization data and information about the surrounding environment with the lightweight \acp{UAV}.
The camera-equipped \ac{UAV} can pass through more confined spaces and cheaply extend the perception range of the \ac{UAV} team due to its low cost.

We propose a novel approach to cooperative flight in \ac{GNSS}-denied environments, providing accurate relative localization between the \acp{UAV}, and enabling direct guidance of less-capable \acp{UAV} by a more-capable \ac{UAV}, mitigating the effects of localization drift.
We focus on the case of a primary \ac{UAV} utilizing 3D \ac{lidar}-based \ac{SLAM} and a lightweight secondary \ac{UAV} carrying a monocular camera utilizing \ac{VIO} (see Fig.~\ref{fig:motivation}).

The proposed approach can enable a variety of multi-UAV tasks in GNSS-denied environments, e.g., cooperative exploration and mapping in confined, sensory-degraded spaces or correcting \ac{VIO} drift of a cooperating team of micro-\acp{UAV}.

% %%}

\subsection{Notations}
    % \vspace{-0.2em}

% NOTATIONS%%{

We denote vectors as bold lowercase letters, matrices as bold upright uppercase letters, and reference frames as simple uppercase letters.
Let $\frames{B}{A}{\vec{t}}$ be the vector describing the position of the origin of frame $A$ in frame $B$.
Let $\frames{B}{A}{\mat{R}} \in SO(3)$ be the rotation matrix from frame $A$ to frame $B$.
We denote
\begin{equation}
  \frames{B}{A}{\mat{T}} = \begin{bmatrix}
    \frames{B}{A}{\mat{R}} & \frames{B}{A}{\vec{t}}\\
    \mat{0}^\mathrm{T} & 1\\
  \end{bmatrix} \in SE(3)
\end{equation}
as the transformation matrix from frame $A$ to frame $B$.
Let $\frames{A}{}{\vec{x}}\tstep{t_k}$ be a 3D position vector in frame $A$ at time $t_k$.
We denote $\frames{A}{}{D}$ as the set of \ac{lidar} detections in reference frame $A$ and $\frames{A}{}{\vec{\chi}}$ as the desired \ac{UAV} trajectory in frame $A$.

% %%}

% PROBLEM STATEMENT%%{

\subsection{Problem statement}
    % \vspace{-0.2em}
\label{sec:problem_statement}
% \vspace{-0.2em}

\begin{figure}
\centering
  \includegraphics[width=1.0\linewidth, trim=0.0cm 0.0cm 0.0cm 0.0cm, clip=true]{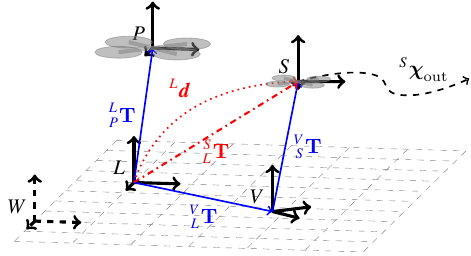}
  \caption{$W$ is the world reference frame.
  Frames $P$ and $S$ correspond to the primary and secondary \ac{UAV}, respectively.
  The primary \ac{UAV} is localized by \ac{lidar} \ac{SLAM} in the local frame $L$, while the secondary \ac{UAV} is localized by \ac{VIO} in the local frame $V$.
  The origins of frames $L,V$ are at the poses where the algorithms were initialized.
  Frames $L$ and $V$ have 4 unobservable \acp{DOF} w.r.t. $W$, resulting in long-term drift.
  Frame $V$ suffers from significantly larger drift than frame $L$.
  $\frames{L}{}{\vec{d}}$ is position of the secondary \ac{UAV} detected from \ac{lidar} data.
  The reference for the secondary \ac{UAV}'s movement is given in the frame $L$, transformed to $\frames{S}{}{\vec{\chi}_\mathrm{out}}$ using the relative localization, and transmitted to the secondary \ac{UAV}.}
\label{fig:drones_tikz}
  % \vspace{-1.5em}
\end{figure}

We focus on tight cooperation in a heterogeneous \ac{UAV} team with various localization sensors.
The crucial part is fusing \ac{lidar} relative localization with \ac{VIO} ego-motion estimates for guidance of a less-capable \ac{UAV} by a more-capable \ac{UAV}.
The more-capable \textit{primary} \ac{UAV} carries a 3D \ac{lidar} and is subject to small long-term localization drift.
The less-capable \textit{secondary} \ac{UAV} is equipped with a monocular camera and suffers from significantly larger long-term localization drift.
Both \acp{UAV} carry an onboard computer, a wireless communication module, a \ac{FCU} with an embedded attitude controller, and an \ac{IMU}.
The \acp{UAV} operate in a \ac{GNSS}-denied environment.
All algorithms run fully on board the \acp{UAV}.

The problem is illustrated in Fig.~\ref{fig:drones_tikz}.
The first task is to estimate the transformation $\frames{V}{L}{\mat{T}}$ between the \ac{lidar} \ac{SLAM} frame $L$ and \ac{VIO} frame $V$ and  the transformation $\frames{S}{L}{\mat{T}}$ between $L$ and the secondary \ac{UAV} body frame $S$ to provide relative localization between the \acp{UAV}.
All reference frames are aligned with the gravity vector obtained from the \ac{IMU} data.
Given general motion, \ac{VIO} always has four unobservable directions, corresponding to the global translation and global yaw.
As reported in~\cite{wuUnobservableDirectionsVINS2016a}, monocular \ac{VIO} additionally exhibits scale unobservability under constant acceleration and global orientation unobservability when the UAV is not rotating.
The scale unobservability under constant acceleration will manifest as long-term drift in the global x, y, z translation of the output.
While the global orientation unobservability given no rotation can cause short-term drift in the direction of the estimated gravity vector, the drift will be corrected once the robot rotates again.
In practice, we assume general rotating motion of the UAV, as the UAV needs to change its tilt even to perform a translational motion, to compensate for wind disturbances while hovering, etc.
Therefore, we assume only negligible short-term drift in the gravity vector with quick corrections back to the correct global values.
Therefore, we estimate 4-\ac{DOF} transformations, consisting of 3D translation and relative heading.

The second task is to use the relative localization data to guide the secondary \ac{UAV} along the desired trajectory $\frames{L}{}{\vec{\chi}}_\mathrm{in}$, defined in the \ac{lidar} \ac{SLAM} reference frame $L$.
The aim is to: 1) give the primary \ac{UAV} the capability to directly guide the secondary \ac{UAV} to, e.g., inspect a specific place or pass around obstacles,
2) mitigate the effects of \ac{VIO} drift to approach the trajectory tracking precision achievable with the less-drifting primary \ac{UAV} localization.

Depending on the environment, the \ac{lidar} relative localization output $\frames{L}{}{\vec{d}}$ is subject to potential false detections and variable delay due to \ac{lidar} data processing time.
The communication between the \acp{UAV} is subject to variable delay depending on the network reliability.
We assume no prior knowledge about the transformation $\frames{V}{L}{\mat{T}}$ or about the initial \ac{UAV} poses.
The proposed approach tries to minimize the computational and payload requirements on the secondary \ac{UAV} by offloading them onto the primary \ac{UAV}.

% %%}

\subsection{Related work}\label{sec:SoA}
    % \vspace{-0.2em}

% RELATED WORK%%{

\subsubsection{UAV relative localization}
In unknown \ac{GNSS}-denied environments, there are several ways to solve relative localization between \acp{UAV}.
Marker-based visual detectors utilize markers, such as \ac{UV} LEDs, in combination with \ac{UV}-sensitive cameras~\cite{walterUVDARSystemVisual2019}.
\bl{Marker-less visual approaches typically rely on a \ac{CNN}~\cite{schillingVisionBasedDroneFlocking2021, vrbaMarkerLessMicroAerial2020a, geVisionbasedRelativeDetection2022, pavlivTrackingRelativeLocalization2021}.
In~\cite{krizekBioinspiredVisualRelative2024}, the authors proposed a visual relative localization method for large \ac{UAV} swarms, estimating the density of \acp{UAV} over distance rather than the positions of individual \acp{UAV}.
However, machine learning-based techniques are susceptible to changes in the appearance of the targets and the environment.
In~\cite{zhangEvDetMAVGeneralizedMAV2025}, the authors proposed a \ac{UAV} detection method from event camera streams based on detecting the spinning propellers.
In~\cite{guoMotionguidedSmallMAV2024}, a motion-guided approach for detecting small \acp{UAV} in complex and non-planar scenes was proposed, combining motion feature enhancement, multi-target tracking, and an appearance-based classifier.
The authors of~\cite{zhengKeypointGuidedEfficientPose2024} proposed a keypoint-guided method for estimating the 6-\ac{DOF} pose of a target \ac{UAV} from monocular camera data, relying on known dimensions of the target \ac{UAV}.
  Although this field of research is progressing, vision-based approaches often struggle under low visibility and fail to provide precise 3D position estimates due to inaccurate distance estimation~\cite{vrbaMarkerLessMicroAerial2020a}, as the approaches require a priori knowledge about the \ac{UAV}'s physical size to estimate the 3D distance from the camera.
Such a priori information may be provided directly or encoded in the neural network using the training dataset.}
\bl{
In~\cite{guoPreservingRelativeLocalization2024}, the authors proposed an active localization correction approach relying on visual detections.
The approach used a \ac{CNN} for detection of cooperating \acp{UAV}, combined with measurements from a depth camera, and focused on controlling the yaw of the \acp{UAV} to balance between obtaining relative localization data and observing the environment.
The limited \ac{FOV} of the cameras is a significant drawback of visual-based methods.
In contrast, many 3D \acp{lidar}, such as the one used in our approach, have full 360$^\circ$ horizontal \ac{FOV}, removing the need to rotate the \ac{UAV} or to use multiple cameras.
Furthermore, in practice, the depth cameras have a usable range of a few meters, compared to tens of meters in the case of the \ac{lidar} sensors.
}

Equipping the \acp{UAV} with \ac{UWB} modules provides distance measurements between pairs of the modules~\cite{guoUltraWidebandOdometryBasedCooperative2020}.
The overall localization accuracy is highly dependent on the shape and size of the \ac{UAV} formation, as it may suffer from unobservabilities and dilution of precision.

A multi-robot \ac{SLAM} method can obtain the relative poses through map merging and inter-robot loop closures, utilizing, e.g., \ac{lidar} data~\cite{huangDiSCoSLAMDistributedScan2022, changLAMPRobustMultiRobot2022a}, stereo camera data~\cite{tianKimeraMultiRobustDistributed2022}, or monocular camera data~\cite{jangMultirobotCollaborativeMonocular2021}.
Such approaches require data of comparable modalities to perform the place recognition / loop closures.
However, \ac{lidar} and visual data use inherently different types of data, as \ac{lidar} data provides information about the geometric structure of the environment, while visual data provides visual features.
Depending on the environment, it may be difficult or even impossible to perform reliable place recognition based on these different sensory modalities. 
Furthermore, such approaches require communicating significant amounts of data or making compromises between the amount of shared data and contained information, influencing the accuracy of the localization~\cite{duboisSharingVisualinertialData2022a}.
Often, they use a ground station for computations, but we aim to run everything on board the computationally-constrained \ac{UAV} platforms.
Our method enables efficient relative localization between heterogeneous UAVs without requiring any place recognition, high-bandwidth data transmission, or ground-station computation.

3D \ac{lidar} data can be used for direct detection of cooperating robots.
In~\cite{grossFieldTestingUAVUGVTeam2019a}, a fusion of 3D \ac{lidar}, fisheye camera, and \ac{UWB} data was used for tracking a \ac{UAV} flying above a \ac{lidar}- and camera-equipped \ac{UGV}, with the \ac{UAV} wirelessly sharing \ac{IMU} and 1D \ac{lidar} data.
In~\cite{zhuSwarmLIODecentralizedSwarm2023, zhuSwarmLIO2DecentralizedEfficient2025}, a decentralized \ac{lidar}-inertial swarm odometry was proposed, utilizing direct detections of cooperating \acp{UAV} from reflectivity values of \ac{lidar} data.
However, these methods tightly couple \ac{lidar} data across the swarm, which prevents heterogeneous configurations and requires heavy, power-demanding, and costly 3D \acp{lidar} on every UAV.
By contrast, our approach enables a primary \ac{lidar}-equipped \ac{UAV} to cooperate with a secondary, lightweight camera-based \ac{UAV}, significantly reducing size, weight, and power requirements for the secondary vehicle.

In this paper, we utilize direct detections of \acp{UAV} from \ac{lidar} data, taking advantage of the robustness of the \ac{lidar} sensor and of the ability to easily obtain precise 3D positions of the cooperating \acp{UAV}.
\bl{The approach is based on our previous work on detection of intruder \acp{UAV} from \ac{lidar} data~\cite{vrbaAutonomousCaptureAgile2022a, vrbaOnboardLiDARbasedFlying2023} and can work without markers or with optional reflective markers aiding the detection process in cluttered environments.
We would like to refer the reader to~\cite{vrbaOnboardLiDARbasedFlying2023} for a thorough analysis of the \ac{lidar}-based detection method, including comparisons with other methods for \ac{UAV} detection.
}

\subsubsection{Cooperative navigation improving UAV localization}
Our approach is similar to methods for improving localization in \ac{GNSS}-challenging environments, where \acp{UAV} with reliable \ac{GNSS}-based localization share data with another \ac{UAV} in a \ac{GNSS}-challenging area~\cite{causaAdaptiveCooperativeNavigation2020, causaCooperativeNavigationVisual2021}.
These approaches utilized visual tracking for relative localization.

Using the relative localization data, we aim to improve the navigation performance of a \ac{UAV} utilizing visual-based self-localization.
Such a goal can be achieved by fusion with \ac{UWB} measurements.
For a single-\ac{UAV} case, a fusion of data from static \ac{UWB} anchors, \ac{lidar} odometry, \ac{IMU}, and \ac{VIO} was proposed in~\cite{nguyenVIRALFusionVisualInertialRangingLidarSensor2022}.
However, placing static \ac{UWB} anchors in the area is not viable in an emergency scenario.
The use of multiple robots, each carrying a \ac{UWB} module, decreases the dependence on external infrastructure.
Such approaches fusing \ac{UWB} and \ac{VIO} were proposed in~\cite{nguyenFlexibleResourceEfficientMultiRobot2022} for collaborative localization of two \acp{UAV} and in~\cite{zieglerDistributedFormationEstimation2021} for distributed formation estimation in large \ac{UAV} swarms.
As previously mentioned, \ac{UWB} modules provide only distance measurements between each pair of devices, and thus the accuracy of such a solution heavily depends on the shape of the \ac{UAV} formation.
In a two-\ac{UAV} team such as ours, \ac{UWB} alone provides only one \ac{DOF} per link, while a single \ac{lidar} detection provides three \acp{DOF}, fully constraining the relative 3D position.

In \cite{xuDecentralizedVisualInertialUWBFusion2020}, the authors fused the detections from a \ac{CNN} with \ac{UWB} data and \ac{VIO} for relative localization in a \ac{UAV} swarm.
The same authors fused map-based inter-\ac{UAV} loop-closures in~\cite{xuOmniSwarmDecentralizedOmnidirectional2022}.
To obtain loop closures, the individual robots need to observe the environment in similar modalities and exchange the data, increasing communication demands.
Thus, loop closures are not suitable for our heterogeneous \ac{UAV} team.
Zhang et al.~\cite{zhangAgileFormationControl2022} fused active visual-based relative localization with \ac{UWB} and \ac{VIO} data for formation control of a marker-equipped \ac{UAV} swarm.
However, such visual-based methods are not robust enough for the target emergency response and \ac{SAR} tasks, which motivate our research.
The authors of~\cite{queraltaVIOUWBBasedCollaborativeLocalization2022} focused on collaborative localization of a \ac{UGV}-\ac{UAV} team.
Their approach relied mainly on \ac{UWB} and \ac{VIO} data, with 3D \ac{lidar} detections utilized during initialization.
In contrast, our approach does not require any \ac{UWB} modules and we utilize the precise 3D positions obtained from the \ac{lidar} data during the entire localization process.

The authors of~\cite{spasojevicActiveCollaborativeLocalization2023} utilized a heterogeneous team of \ac{lidar}-carrying \acp{UGV} and camera-equipped \acp{UAV} with the goal of detecting the \acp{UGV} from onboard cameras on \acp{UAV} and using them as landmarks for improving the \ac{UAV} localization.
Their work focused mainly on the optimal placement of the \acp{UGV} in the environment.
In contrast, we focus on a heterogeneous \ac{UAV}-only team with the \ac{lidar}-carrying \ac{UAV} detecting the camera-equipped \ac{UAV}.
Relative localization in a \ac{UAV} team requires obtaining their 3D positions very accurately and 3D \ac{lidar} sensors represent an ideal choice for such a task.

\subsubsection{VIO drift mitigation}
In~\cite{yuMonocularCameraLocalization2020}, \ac{VIO} drift was mitigated by finding feature correspondences in \ac{VIO} data and prior \ac{lidar} maps.
Our approach does not require prior maps and can be deployed in previously unseen environments.
In \ac{GNSS}-enabled environments, \ac{VIO} drift can be corrected by fusing raw Doppler velocities and pseudoranges~\cite{caoGVINSTightlyCoupled2022} or 3D \ac{GNSS} positions~\cite{leeIntermittentGPSaidedVIO2020}.
  In~\cite{leeIntermittentGPSaidedVIO2020}, the authors obtained the \ac{GNSS}-\ac{VIO} frame transformation using window-based optimization and transformed the \ac{VIO} state to the \ac{GNSS} frame during initialization to avoid unobservabilities.
  The \ac{GNSS} data are in the global frame but suffer from short-term meter-level drift.
  \ac{lidar} data are in a local frame, resulting in minor long-term drift w.r.t. global frame, but exhibit centimeter-level precision.
  Our approach uses window-based optimization not only for initialization but during the entire estimation process.

In~\cite{eckenhoffTightlyCoupledVisualInertialLocalization2019,eckenhoffSchmidtEKFbasedVisualInertialMoving2020}, the authors focused on \ac{VIO} and visual target tracking, improving the performance of both thanks to their tight coupling.
We instead improve the navigation performance of a \ac{VIO}-using \ac{UAV} by guiding it by another \ac{UAV} with less-drifting localization.
We opted for a loosely-coupled approach to mitigate issues with modeling errors hurting the performance, which can be more pronounced in a tightly-coupled estimator~\cite{eckenhoffSchmidtEKFbasedVisualInertialMoving2020}, and to enable our approach to be universally usable with any \ac{VIO} algorithm.

\subsubsection{Contributions}
The proposed relative localization method can accurately obtain 3D positions of \acp{UAV} without prior knowledge of their size.
It is robust to low illumination due to the nature of the \ac{lidar} sensor and does not require any additional hardware apart from the 3D \ac{lidar} that is simultaneously used for the ego-localization of the primary \ac{UAV}.
Our approach poses minimal requirements on the secondary \ac{UAV}, making the method compatible with a number of commercially-available \acp{UAV} and state-of-the-art \ac{VIO} algorithms.
It requires no prior information about the initial \ac{UAV} poses.
It is worth mentioning that the proposed approach is not limited to the use of \ac{VIO} algorithms on board the secondary \ac{UAV}, but can be easily applied to any odometry method providing localization data in some local reference frame thanks to its loosely-coupled nature.

The approach proposed in this paper is an extension of our previous work on utilizing 3D \ac{lidar} detections to guide a cooperating \ac{UAV}~\cite{pritzl2022icuas}.
In this paper, we extend the relative localization approach by fusing the \ac{lidar} detections with the \ac{VIO} output, thus obtaining relative orientation between the \acp{UAV} and enabling to guide the secondary \ac{UAV} along target trajectories with a variable desired heading.
Simultaneously, the fusion approach removes the need for any a~priori knowledge about the \ac{UAV} poses.
The proposed method has been used to enable subsequent research~\cite{pritzlCooperativeFlight2024} on cooperative guiding of drones in cluttered environments~\cite{pritzlDronesGuidingDrones2024} and cooperative indoor exploration~\cite{cihlarovaCooperativeIndoor2024}.
Note that the aforementioned publications do not focus on the fusion of \ac{VIO} with \ac{lidar} relative localization but utilize the method proposed in this paper for higher-level tasks enabled by this approach.
The short workshop paper~\cite{pritzlCooperativeFlight2024} provides only a high-level summary of these methods and their uses, without any details.

To the best of our knowledge, the proposed approach is the first solution to relative localization between heterogeneous \acp{UAV} based on \ac{lidar} and \ac{VIO} fusion.
The key contribution of our work is to fuse the complementary strengths of \ac{lidar} (accuracy, robustness to environmental conditions) and cameras (low weight, size, cost, and power demand) in a single heterogeneous \ac{UAV} team. 
As shown in the experimental evaluation, such a setup provides accurate localization between the \acp{UAV} and enables precise guidance of the secondary \ac{UAV} with drifting \ac{VIO}.
  In a practical application, the proposed approach can be employed for guiding less-capable \acp{UAV} with minimal sensory payload through areas of visual sensory degradation.

% %%}

The contributions of this work can be summarized as:
\begin{itemize}
  \item A novel approach to \ac{UAV} relative localization based on the fusion of \ac{lidar} and \ac{VIO} ego-motion data, which provides a common reference frame for a team of heterogeneously localized \acp{UAV}.
    Unlike existing solutions that rely on homogeneous sensors or lower-accuracy modalities, our approach leverages the high precision of 3D \ac{lidar} to achieve accurate relative localization.
    The proposed method enables accurate estimation of the time-varying relative transformation between the different reference frames of the \acp{UAV}, and is able to quickly react to the coordinate frame drift while taking into account possible false detections and measurement delays.
  \item A cooperative guidance strategy utilizing the relative localization that allows the secondary \ac{UAV} to track desired trajectories defined in the primary \ac{UAV} reference frame while mitigating \ac{VIO} drift.
  \item The raw data from our real-world experiments have been released online to provide a unique dataset useful for experimental verification of such relative localization approaches to the research community.
\end{itemize}

%%{ SYSTEM MODEL

\section{Cooperative localization and guidance approach}
    % \vspace{-0.3em}
\label{sec:system_model}
\begin{figure*}
  \centering
  % \input{figures/diagram.tex}
  % \resizebox{1.0\linewidth}{!}{\input{figures/diagram.tex}}
  \includegraphics[width=1.0\linewidth, trim=0.0cm 0.0cm 0.0cm 0.0cm, clip=true]{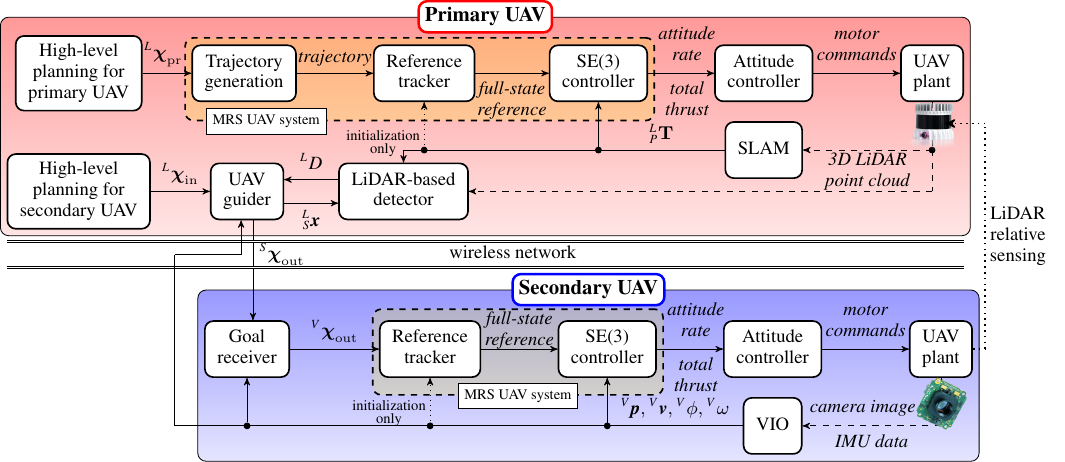}
  \caption{The primary \ac{UAV} tracks the pose of the secondary \ac{UAV} from the fusion of \ac{lidar} and \ac{VIO} data and transmits commands to the secondary \ac{UAV} over a wireless network. Movement of the \ac{UAV} pair is planned on board the primary \ac{UAV}. Position control in the frame of each respective self-localization method is provided by the MRS UAV System~\cite{bacaMRSUAVSystem2021}.}
  \label{fig:diagram}
  % \vspace{-1.5em}
  % \vspace{-1.2em}
\end{figure*}

The overall architecture of the proposed approach (see Fig.~\ref{fig:diagram}) is based upon the results of our preliminary work~\cite{pritzl2022icuas}, which assessed the feasibility of using raw \ac{lidar} detections to guide a cooperating \ac{UAV} to desired positions.
The \ac{UAV} pair forms a closed feedback loop controlling the secondary \ac{UAV} pose in the frame $L$.
High-level planning and estimation run on board the primary \ac{UAV}.
The secondary \ac{UAV} receives the desired trajectories $\frames{S}{}{\vec{\chi}}_\mathrm{out}$ and tracks references in its local frame $V$.
The \ac{UAV} pair is time-synchronized over the wireless network.
This paper focuses mainly on the \textit{UAV guider} module (see Fig.~\ref{fig:diagram_uav_guider}).
The control pipeline used for tracking references in the local frames of each self-localization method is provided by the MRS UAV System~\cite{bacaMRSUAVSystem2021}\footnote{\url{https://github.com/ctu-mrs/mrs_uav_system}}.

\subsection{LiDAR-based detector}
    % \vspace{-0.2em}

The \textit{\ac{lidar}-based detector} performs multi-target detection and tracking of flying objects in the 3D \ac{lidar} point clouds in frame $L$.
The \ac{lidar} data processing is based on our work on autonomous interception of intruder \acp{UAV}~\cite{vrbaAutonomousCaptureAgile2022a, vrbaOnboardLiDARbasedFlying2023}.

The detector constructs a voxel-based occupancy map of the surrounding environment from \ac{lidar} data and detects flying objects in the map.
A Kalman filter-based point cluster-tracking approach is utilized to compensate for the uncertainty in the sensory data and mitigate delays caused by the processing time of the detection algorithm.
Here, we describe the differences employed for its use in the proposed cooperative localization and guidance approach and encourage the reader to look into~\cite{vrbaOnboardLiDARbasedFlying2023} for more details about the \textit{\ac{lidar}-based detector} itself, including its extensive experimental evaluation.

The \textit{\ac{lidar}-based detector} additionally receives the estimated \ac{UAV} position $\frames{L}{S}{\vec{x}}$ from the \textit{\ac{UAV} guider} module.
The position is used in cluster-to-target association instead of the tracked target position if $\frames{L}{S}{\vec{x}}$ is located within a desired uncertainty radius from a tracked target. 
The position $\frames{L}{S}{\vec{x}}$ can be more accurate due to the incorporation of the low-delay \ac{VIO} data.

\bl{The detected \ac{UAV} can be optionally equipped with reflective markers to improve the detection robustness in cluttered environments.
If the reflective markers are used, the detector processes reflectivity information contained in the \ac{lidar} point cloud and discards point clusters that do not contain points above a predefined reflectivity threshold.
When the reflective markers are used, the accuracy of the detections themselves stays the same, as the detections are produced by calculating the centroids of the original LiDAR clusters.
The same LiDAR points would be used in the marker-less variation of the method.
The reflective markers are passive.
They only reflect incoming LiDAR beams and do not require any power source on board the UAV, unlike, e.g., the UV-based markers of~\cite{walterUVDARSystemVisual2019}.
The advantages of the use of reflective markers are:
\begin{itemize}
    \item They provide an additional safety feature in cluttered environments.
      Although their use is not strictly necessary, it may be a desirable safety feature depending on the objects in the environment.
      For example, imagine a ceiling light hanging on a thin cable.
      Depending on the parameters of the LiDAR and the distance from the light, the LiDAR may not observe the cable in the LiDAR data, and the light will therefore appear to be a flying object, easily mistaken for a UAV.
      If the actual secondary UAV flies close to the ceiling light, the false detection may be falsely associated with the UAV, especially in the presence of \ac{VIO} drift.
\end{itemize}
Disadvantages of the reflective markers:
\begin{itemize}
  \item The reflective markers need to be placed on the UAV, requiring a slight modification of the UAV in comparison to the marker-less method.
    Since we assume cooperating \acp{UAV} (unlike in~\cite{vrbaOnboardLiDARbasedFlying2023}), this is not a limitation of the proposed approach.
    \item If the reflective marker is too small and is not observed in the LiDAR data, correct UAV detections may be discarded.
      Such a situation may happen depending on the parameters of the LiDAR, i.e., on the spread between the LiDAR beams, on the distance between the UAVs, and on the size of the reflective markers.
\end{itemize}
Based on the presented advantages and disadvantages, the reflective markers are not necessary in open areas, but are an important safety feature in cluttered environments.

The detected 3D positions of all tracked objects are sent to the \textit{\ac{UAV} guider module} in the set of detections $\frames{L}{}{D}$.
}

\subsection{UAV guider}
\begin{figure}
  \centering
  \includegraphics[width=1.0\linewidth, trim=0.0cm 0.0cm 0.0cm 0.0cm, clip=true]{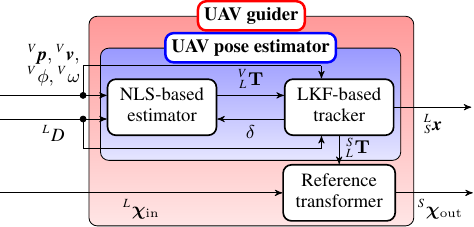}
  \caption{The UAV guider module fuses \ac{lidar} detections with \ac{VIO} data to estimate 4-DOF poses of the secondary \ac{UAV}.
  The estimation result is used for transformation of references to the coordinate frame of the secondary \ac{UAV}.}
  \label{fig:diagram_uav_guider}
  % \vspace{-1.5em}
  % \vspace{-1.2em}
\end{figure}

    % \vspace{-0.2em}
The \textit{\ac{UAV} guider} (see Fig.~\ref{fig:diagram_uav_guider}) estimates the transformation $\frames{S}{L}{\mat{T}}$ from the \ac{lidar} detections and \ac{VIO} data, and transforms the desired trajectories to the secondary \ac{UAV}'s frame $S$.
  The \ac{UAV} pose is estimated in two loosely-coupled steps.
  The \textit{\ac{NLS}-based estimator} aligns \ac{lidar} detections and \ac{VIO} poses in a sliding window to produce $\frames{V}{L}{\mat{T}}$.
  The transformation $\frames{V}{L}{\mat{T}}$ connects the reference frames but lags behind the true transformation values in areas with large \ac{VIO} drift.
  Therefore, the \textit{\ac{LKF}-based tracker} utilizes $\frames{V}{L}{\mat{T}}$ and fuses the latest \ac{lidar} detections and \ac{VIO} to provide a drift-compensated estimate of $\frames{S}{L}{\mat{T}}$ usable for transformation of the desired trajectories.
  The \textit{\ac{LKF}-based tracker} also deals with measurement delays and provides association of \ac{lidar} detections based on their Mahalanobis distance.

\subsubsection{NLS relative transformation optimization}\label{sec:nls}
The 4-\ac{DOF} transformation between the \ac{lidar} \ac{SLAM} frame $L$ and the \ac{VIO} frame $V$ is defined as
\begin{equation}
  \frames{V}{L}{\mat{T}} = \begin{bmatrix}
    \frames{V}{L}{\mat{R}}(\theta) & \frames{V}{L}{\vec{t}}\\
    \mat{0}^\mathrm{T} & 1\\
  \end{bmatrix} \in SE(3).
\end{equation}
The problem of estimating its parameters from a set of \ac{lidar}-\ac{VIO} pose correspondences in a sliding window can be formulated as
\begin{equation}\label{eqn:nls}
  % \footnotesize
  \frames{V}{L}{\hat{\vec{t}}}, \hat{\theta}= \arg\min_{\frames{V}{L}{\vec{t}}, \theta} \frac{1}{2}\sum_i\rho\left(\left|\left|\frames{V}{L}{\mat{R}}(\theta)\frames{L}{}{\vec{d}}_{\tstep{t_i}} + \frames{V}{L}{\vec{t}} - \frames{V}{}{\vec{p}}_{\tstep{t_i}}\right|\right|^2\right),
\end{equation}
\begin{equation}
  % \footnotesize
  \frames{V}{L}{\vec{t}} \in \mathbb{R}^3,~\theta\in [-\pi,\pi],
\end{equation}
where $\frames{V}{L}{\vec{t}}$ is the relative translation vector between the reference frames, $\theta$ is the relative heading, $\frames{L}{}{\vec{d}}_{\tstep{t_i}}$ is the \ac{lidar} detection position at time $t_i$, and $\frames{V}{}{\vec{p}}_{\tstep{t_i}}$ is the corresponding \ac{VIO} position.
$\rho()$ denotes the loss function used for reducing the influence of outliers on the optimization solution.
In the experimental evaluation of the proposed approach, the Levenberg-Marquardt algorithm, implemented in the Ceres Solver~\cite{AgarwalCeresSolver2022}, was employed for solving the \ac{NLS} problem.
The soft L1 loss function, defined as
\begin{equation}
  \rho(s) = 2\left(\sqrt{1+s}-1\right),
\end{equation}
was employed, and the automatic differentiation functionality of the Ceres Solver was used to solve the problem.

The length of the sliding window is constant and predefined.
The formulation of the optimization problem assumes constant $\frames{V}{L}{\mat{T}}$ over the course of the sliding window.
  The \ac{lidar} detections exhibit high accuracy with little long-term drift.
  The \ac{VIO} positions exhibit significant long-term drift, especially in texture-less environments.
  Therefore, a longer window does not necessarily equal a better estimate.
  The window length must balance between measurement noise mitigation and long-term drift tracking.
  The optimal length selection may change with the environment, e.g., in an environment with a lot of features and little \ac{VIO} drift, the window can be longer, whereas in a featureless environment with significant \ac{VIO} drift, a shorter window is desired.

The \ac{lidar}-\ac{VIO} correspondence set $\left\{\left[\frames{L}{}{\vec{d}}\tstep{t_i}, \frames{V}{}{\vec{p}}\tstep{t_i} \right]_i\right\}$ is constructed from the \ac{VIO} position buffer by linearly interpolating the \ac{lidar} detections to time stamps corresponding to each \ac{VIO} position in the buffer.
The result of the \ac{NLS} optimization is only used if the optimization converges and its final cost is lower than a predefined threshold.

\subsubsection{LKF-based tracker}
The \ac{LKF}-based tracker predicts the state of the secondary \ac{UAV} in the frame $L$ and outputs the estimated \ac{UAV} position $\frames{L}{S}{\vec{x}}$ and matrix $\frames{S}{L}{\mat{T}}$ used for transforming the desired references for secondary \ac{UAV}.

The tracker is based on an \ac{LKF} with a recalculating history buffer~\cite{pritzlAdaptiveAltitudeEstimation}.
The history buffer contains recent measurements and measurement covariance matrices sorted by time stamps.
When the tracker predicts its state at a specific time, it is calculated using information from the oldest element in the buffer up to the desired time stamp.
The tracker is based upon a 4-\ac{DOF} constant-velocity model with a state vector
\begin{equation}
  \vec{x}^\mathrm{KF} = \begin{bmatrix}
    \frames{L}{}{\vec{x}}^{\mathrm{T}} & \frames{L}{}{\vec{v}}^\mathrm{T} & \frames{L}{}{\phi} & \frames{L}{}{\omega}
  \end{bmatrix}^\mathrm{T}_{8\times1},
\end{equation}
where $\frames{L}{}{\vec{x}}$ is the 3D position of the secondary \ac{UAV} in the \ac{lidar} \ac{SLAM} frame, $\frames{L}{}{\vec{v}}$ is the velocity of the \ac{UAV}, $\frames{L}{}{\phi}$ is the heading, and $\frames{L}{}{\omega}$ is the heading rate.

The state covariance matrix can be expressed as
\begin{equation}
  \mat{\Sigma}^
  \mathrm{KF}= \begin{bmatrix}
    \mat{\Sigma}_{xx} & \mat{\Sigma}_{xv} & 0 & 0 \\
    \mat{\Sigma}_{vx} & \mat{\Sigma}_{vv} & 0 & 0 \\
    0  &0   & \Sigma_{\phi \phi} & \Sigma_{\phi \omega} \\
    0  & 0  & \Sigma_{\omega \phi} & \Sigma_{\omega \omega} \\
  \end{bmatrix}.
\end{equation}
To achieve consistent performance when fusing unsynchronized measurements, the state covariance matrix is propagated forward in time at each prediction step using a continuous-time representation with zero-mean white noise on linear velocity and heading rate~\cite{eade2017}.
The linear position and velocity submatrices from time $t_k$ are propagated to time $t_{k+1}$ based on the noise vector \vec{$\sigma}^2_v$ as
\begin{equation}
  \mat{\Sigma}_{vv[t_{k+1}]} = \mat{\Sigma}_{vv[t_{k}]} +  \vec{\sigma}^2_v \Delta t
\end{equation}
\begin{equation}
  \mat{\Sigma}_{xv[t_{k+1}]} = \mat{\Sigma}_{xv[t_{k}]} +  \mat{\Sigma}_{vv[t_{k}]} \Delta t + \vec{\sigma}^2_v \frac{\Delta t^2}{2} 
\end{equation}
\begin{equation}
  \mat{\Sigma}_{vx[t_{k+1}]} = \mat{\Sigma}_{vx[t_{k}]} +  \mat{\Sigma}_{vv[t_{k}]} \Delta t + \vec{\sigma}^2_v \frac{\Delta t^2}{2} 
\end{equation}
\begin{equation}
  \mat{\Sigma}_{xx[t_{k+1}]} = \mat{\Sigma}_{xx[t_{k}]} +  2\mat{\Sigma}_{xv[t_{k}]} \Delta t + \mat{\Sigma}_{vv[t_{k}]} \Delta t^2 + \vec{\sigma}^2_v \frac{\Delta t^3}{3},
\end{equation}
where
\begin{equation}
 \Delta t = t_{k+1} - t_k.
\end{equation}
The heading and heading rate submatrices are updated analogously using the heading rate noise variance $\sigma^2_{\omega}$.
The linear velocity noise is assumed to be isotropic, thus the process noise depends only on the parameters $\sigma^2_v$ and $\sigma^2_\omega$.
% \textbf{TODO: process noise} \cite{eade2017}

The measurement vectors used for updating the estimate are
\begin{equation}\label{eqn:measurements}
  \vec{z}^\mathrm{KF}_{\mathrm{L}} = \begin{bmatrix}
    \frames{L}{}{\vec{d}}
  \end{bmatrix}_{3\times1},~
  \vec{z}^\mathrm{KF}_\mathrm{V} = \begin{bmatrix} \vec{z}^{\mathrm{T}}_{x} & \vec{z}^{\mathrm{T}}_v & z_\phi & z_\omega
  \end{bmatrix}^\mathrm{T}_{8\times1},
\end{equation}
  where $\vec{z}^\mathrm{KF}_\mathrm{L}$ is constructed each time a new \ac{lidar} detection $\frames{L}{}{\vec{d}}$ is associated to the estimate $\vec{x}^\mathrm{KF}$.
  If the latest \ac{VIO} pose from time $t_l$ is newer than the latest \ac{lidar} measurement from time $t_k$, the measurement vector $\vec{z}^\mathrm{KF}_\mathrm{V}$ is used, with the position measurement calculated as
\begin{equation}\label{eqn:vio_pos}
  \vec{z}_{\mathrm{x}\tstep{t_l}} = \frames{L}{}{\vec{d}}\tstep{t_k} + \frames{L}{V}{\mat{R}}(\theta)\left(
  \frames{V}{}{\vec{p}}\tstep{t_l} - \frames{V}{}{\vec{p}}\tstep{t_k}
  \right),
\end{equation}
  where $\frames{L}{}{\vec{d}}\tstep{t_k}$ is the latest \ac{lidar} detection, $\frames{V}{}{\vec{p}}\tstep{t_l}$ is the new \ac{VIO} pose, $\frames{V}{}{\vec{p}}\tstep{t_k}$ is the \ac{VIO} pose interpolated to time $t_k$, and $\frames{L}{V}{\mat{R}}(\theta)$ is the rotation matrix constructed from the latest \ac{NLS} optimization output.
The velocity measurement is calculated as
\begin{equation}\label{eqn:vio_velocity}
  {\vec{z}}_{v\tstep{t_l}} = \frames{L}{V}{\mat{R}}(\theta) \frames{V}{}{\vec{v}}\tstep{t_l},
\end{equation}
where $\frames{V}{}{\vec{v}}_{\tstep{t_l}}$ is the velocity obtained from the \ac{VIO}.
The heading and heading rate measurements are constructed as
\begin{equation}\label{eqn:headings}
  z_{\phi\tstep{t_l}} = \frames{V}{}{\phi\tstep{t_l}} - \theta,~
  z_{\omega\tstep{t_l}} = \frames{V}{}{\omega\tstep{t_l}},
\end{equation}
where $\frames{V}{}{\phi\tstep{t_l}}$ is the heading of the \ac{VIO} pose, $\theta$ is the relative heading obtained from the \ac{NLS} optimization, and $\frames{V}{}{\omega\tstep{t_l}}$ is the heading rate from the \ac{VIO} output.
If the received \ac{VIO} pose is older than the latest \ac{lidar} measurement, only the heading and heading rate states are updated with the measurement vector constructed as
\begin{equation}
  \vec{z}^\mathrm{KF}_\phi = \begin{bmatrix} z_\phi & z_\omega
  \end{bmatrix}^\mathrm{T}_{2\times 1},
\end{equation}
where its elements are calculated by (\ref{eqn:headings}).

Measurement covariance matrix $\mat{R}^\mathrm{KF}_\mathrm{L}$ is constructed as
\begin{equation}
  \mat{R}^\mathrm{KF}_\mathrm{L} = \begin{bmatrix}
    \mat{R}^\mathrm{KF}_{\vec{x}} & \mat{0} & 0 & 0 \\
    \mat{0} & \mat{R}^\mathrm{KF}_{\vec{v}} & 0 & 0 \\
    0 & 0 &  \mathrm{R}^\mathrm{KF}_{\phi} & 0 \\
    0 & 0 & 0 & \mathrm{R}^\mathrm{KF}_{\omega} \\
  \end{bmatrix},
\end{equation}
where the individual submatrices $\mat{R}^\mathrm{KF}_{\vec{a}}$ are obtained by propagating the uncertainty of the individual measurements from equations~(\ref{eqn:vio_pos}), (\ref{eqn:vio_velocity}), and (\ref{eqn:headings}) as
\begin{equation}
  \mat{R}^\mathrm{KF}_{\vec{a}} = \mat{J}_{\vec{a}} \mat{R}^\mathrm{orig}_{\vec{a}} \mat{J}_{\vec{a}}^\mathrm{T},
\end{equation}
where $\mat{J}_{\vec{a}}$ is the Jacobian matrix of the respective measurement functions~(\ref{eqn:vio_pos}), (\ref{eqn:vio_velocity}), or (\ref{eqn:headings}), and $\mat{R}^\mathrm{orig}_{\vec{a}}$ is the covariance matrix of the original measurements.
The measurement noise is assumed to be isotropic, thus the measurement covariance matrices depend on the detection variance $\sigma^2_d$, relative heading variance $\sigma^2_\theta$, VIO relative vector variance $\sigma^2_r$, VIO linear velocity variance $\sigma^2_V$, VIO heading variance $\sigma^2_\phi$, and VIO heading rate variance $\sigma^2_\omega$.

  New \ac{lidar} detections are associated to the estimate $\vec{x}^\mathrm{KF}$ based on their Euclidean and Mahalanobis distance.
If the Euclidean distance of a detection from the estimate is lower than a predefined threshold, its squared Mahalanobis distance from the estimate at the corresponding time is calculated as
\begin{equation}
  \delta^2\tstep{t_k} = \vec{y}\tstep{t_k}^\mathrm{T}\mat{S}^{-1}\tstep{t_k}\vec{y}\tstep{t_k},
\end{equation}
where $\vec{y}\tstep{t_k}$ is the measurement innovation and $\mat{S}\tstep{t_k}$ is the innovation covariance calculated during the \ac{LKF} correction.
The detection with the lowest Mahalanobis distance $\delta\tstep{t_k}$ is selected and associated to the estimate if it is lower than a predefined threshold.

\subsubsection{Estimate initialization}
\label{sec:initialization}

  The \textit{\ac{lidar}-based detector} provides a set of detections $\frames{L}{}{D}$ of all the tracked flying objects in the surrounding environment.
  The \textit{\ac{UAV} pose estimator} constructs a separate detection buffer for each tracked object and performs the \ac{NLS} trajectory alignment (\ref{eqn:nls}) for each detection buffer and the \ac{VIO} pose buffer.
  If the buffers contain sufficient movement of the \ac{UAV} and the \ac{NLS} optimization converges to a solution with the final cost below a specified threshold, the estimate is initialized based on this detection buffer.

\subsubsection{Reference transformer}

The \textit{Reference transformer} utilizes the relative transformation $\prescript{S}{L}{\mat{T}}$ produced by the estimator to transform the desired trajectory $\frames{L}{}{\vec{\chi}}_{\mathrm{in}}$ to the secondary \ac{UAV} reference frame $S$.
The reference transformer keeps track of what part of the desired trajectory has already been completed in order to transmit only unvisited parts and repeatedly transmits references to the secondary \ac{UAV} at the rate of \SI{5}{Hz}.
On board the secondary \ac{UAV}, the \textit{Goal receiver} module receives the trajectories and checks them for time stamp inconsistencies before transferring them to the secondary \ac{UAV} control pipeline.

\subsection{Analysis of degenerate cases}
    % \vspace{-0.2em}
In the proposed approach, these degenerate cases can arise:
  \subsubsection{Loss of LiDAR detections \texorpdfstring{$\frames{L}{}{\vec{d}}$}{TEXT}}
    When line-of-sight visibility between the \acp{UAV} gets broken, the transformations $\frames{S}{L}{\mat{T}}$ and $\frames{V}{L}{\mat{T}}$ become unobservable.
    The estimator will keep tracking the UAV from the \ac{VIO} poses transformed using the latest $\frames{V}{L}{\mat{T}}$ (see (\ref{eqn:measurements})-(\ref{eqn:headings})), but the estimated \ac{UAV} pose in the frame $L$ will drift away from its true value, as $\frames{V}{L}{\mat{T}}$ will not be getting updated.
    When visibility between the \acp{UAV} gets re-established, the estimate will be corrected or re-initialized.

  \subsubsection{Loss of VIO poses \texorpdfstring{$\frames{V}{}{\vec{p}}$}{TEXT}}
    When communication between the \acp{UAV} gets broken, its 3D position $\frames{L}{S}{\vec{x}}$ is still observable from the \ac{lidar} data, but the transformation $\frames{V}{L}{\mat{T}}$ and the heading $\frames{L}{}{\phi}$ of the secondary UAV become unobservable.
    The estimator will keep tracking the \ac{UAV} but will not update the $\frames{L}{}{\phi}$ and $\frames{L}{}{\omega}$ states during the loss of communication.

  \subsubsection{Small translational motion of the secondary \ac{UAV}}
    The relative transformation $\frames{V}{L}{\mat{T}}$ becomes unobservable as the \ac{UAV} trajectory in the sliding window degrades to a single point.
    If the observed \ac{UAV} motion falls to the level of the \ac{lidar} measurement noise, the resulting estimate will become inconsistent.
    Such a situation is detected based on the trajectory length in the window and the eigenvalues of the Fisher information matrix
\begin{equation}
  \mat{F}\left(\frames{V}{L}{\hat{\vec{t}}}, \hat{\theta}\right) = \mat{J}\left(\frames{V}{L}{\hat{\vec{t}}}, \hat{\theta}\right)^\mathrm{T} \mat{J}\left(\frames{V}{L}{\hat{\vec{t}}}, \hat{\theta}\right),
\end{equation}
    constructed from the Jacobian $\mat{J}$ evaluated when solving the \ac{NLS} problem.
    As the optimization problem becomes degenerate, the lowest eigenvalue of the matrix $\mat{F}$ tends to zero.
    If the eigenvalue falls below a predefined threshold, the estimator stops updating $\frames{V}{L}{\mat{T}}$.

%%}

%%{ EXPERIMENTS

\section{Experimental verification}

Video of the experimental verification is available online~\footnote{\url{http://mrs.felk.cvut.cz/coop-fusion}}.

\label{sec:experiments}
    % \vspace{-0.3em}

% \begin{figure}
%   \centering
%   \begin{tikzpicture}
%     \node[anchor=north west,inner sep=0,draw=black] (a) at (0, 0)
%     {
%       \includegraphics[width=0.48\linewidth, trim=3cm 2cm 3cm 2cm, clip=true]{figures/x500.jpg}
%     };
%     \node[fill=white,draw=black,text=black, anchor=north west] at (a.north west) {\footnotesize (a)};

%     \node[anchor=north west,inner sep=0,draw=black] (b) at (4.4cm, 0cm)
%     {
%       \includegraphics[width=0.48\linewidth, trim=0.0cm 0.0cm 0.0cm 0.0cm, clip=true]{figures/vio_drona4.jpg}
%     };
%     \node[fill=white,draw=black,text=black, anchor=north west] at (b.north west) {\footnotesize (b)};

%   \end{tikzpicture}
%   \caption{The \ac{UAV} platforms employed in the real-world experiments: (a) Primary \ac{UAV} carrying the Ouster 3D \ac{lidar}, (b) Secondary \ac{UAV} with the monocular camera.%
%   }
%   \label{fig:drona}
% \end{figure}

\subsection{Simulations}
    % \vspace{-0.2em}

\subsubsection{Drift compensation}
\label{sec:sim_drift_compensation}
\begin{figure}[t]
  \centering
  \includegraphics[width=1.0\linewidth, trim=0.0cm 0.0cm 0cm 0.81cm, clip=true]{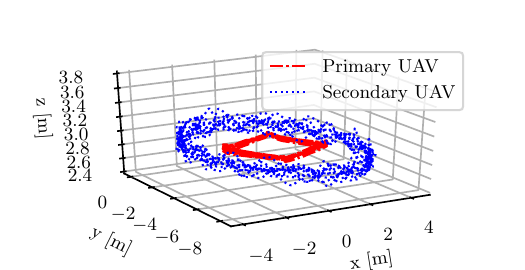}
  \caption{Trajectories in the simulation with zero drift.}
  \label{fig:sim_3d_plot}
  % \vspace{-1.0em}
\end{figure}

Simulations in the Gazebo simulator were performed to evaluate the ability to compensate for the odometry drift.
The simulations were performed in an indoor environment with no obstacles breaking the line of sight between the \acp{UAV} (see the multimedia attachment).
In the simulated scenario, the primary \ac{UAV} moved in a 3 by 3 \si{m} large square while guiding the secondary \ac{UAV} to follow a surrounding circular trajectory with the radius of \SI{4}{m} at the velocity of \SI{0.5}{\meter\per\second} (see Fig.~\ref{fig:sim_3d_plot}).
The trajectories were predefined in the \ac{lidar} \ac{SLAM} reference frame.
In each run of the simulation, the secondary \ac{UAV} was tasked to complete the circle 10 times.

The primary \ac{UAV} carried a simulated \ac{lidar} with the same parameters as its real-world counterpart from Sec.~\ref{sec:realworld} and utilized it in the LOAM~\cite{zhangLOAMLidarOdometry2014} \ac{SLAM} algorithm.
Instead of real \ac{VIO}, the secondary \ac{UAV} control loop used ground-truth localization with artificially inserted constant-velocity drift in the $x$-axis.
The approach was evaluated for drift values ranging from 0 to 1~\si{\meter\per\second} with the step of \SI{0.1}{\meter\per\second}.
For each drift value, the simulation was done 10 times.
The simulation run was considered to be a failure, if the secondary \ac{UAV} failed to complete the 10 circles, e.g., due to collision with the surrounding obstacles.

\begin{figure}[t]
  \centering
  \includegraphics[width=1.0\linewidth, trim=0.0cm 0.35cm 0cm 0.35cm, clip=true]{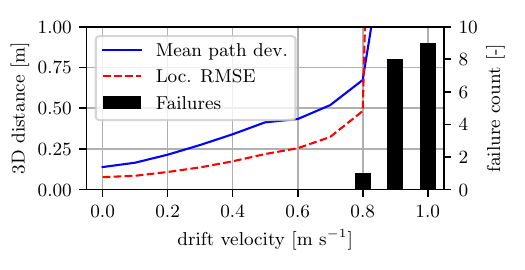}
  \caption{Mean deviation from the circular path, relative localization \ac{RMSE}, and failure count for each drift value.}
  \label{fig:sim_deviation}
  % \vspace{-1.0em}
\end{figure}

Fig.~\ref{fig:sim_deviation} shows the mean deviation from the circular path, the \ac{RMSE} of the relative localization, and the number of runs in which the \ac{UAV} failed to complete the 10 circles.
The simulations were completely successful for drift values up to \SI{0.7}{\meter\per\second} and with one failure for the drift of \SI{0.8}{\meter\per\second}.
The mean path deviation ranged from \SI{0.14}{\meter} in the case of zero drift up to \SI{0.67}{\meter} in the case of \SI{0.8}{\meter\per\second} drift.
The relative localization \ac{RMSE} ranged from \SI{0.08}{\meter} up to \SI{0.48}{\meter}.

  \subsubsection{\ac{NLOS} simulation}
  \label{sec:sim_nlos}
\begin{figure}[t]
  \centering
  \includegraphics[width=1.0\linewidth, trim=0.3cm 0.35cm 0.2cm 0.3cm, clip=true]{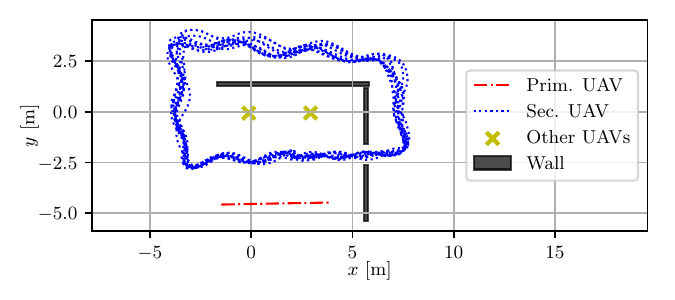}
  \caption{Ground-truth trajectories from \ac{NLOS} simulation.}
  \label{fig:sim_nlos_gt}
  % \vspace{-1.0em}
\end{figure}

The approach was evaluated in a simulated \ac{NLOS} situation.
The primary \ac{UAV} moved back and forth in a line and guided the secondary \ac{UAV} to follow a trajectory passing through a \SI{1}{m} wide gap and behind a wall (see Fig.~\ref{fig:sim_nlos_gt}).
The secondary \ac{UAV} was \SI{0.45}{m}$~\times~$\SI{0.45}{m} wide including propellers.
The secondary \ac{UAV} performed the loop 10 times.
Two additional \acp{UAV} hovered next to the desired trajectory to demonstrate the robustness of the approach to false detections.
As in the previous simulation, the primary \ac{UAV} carried a simulated Ouster OS0-128 3D \ac{lidar} with the \ac{lidar} data utilized by the LOAM~\cite{zhangLOAMLidarOdometry2014} SLAM.
The secondary \ac{UAV} carried a fisheye camera with \ac{IMU} corresponding to those used in the real-world experiments and utilized VINS-Mono~\cite{qinVINSMonoRobustVersatile2018} in its control loop.

\begin{figure}[t]
  \centering
  \includegraphics[width=1.0\linewidth, trim=0.4cm 0.35cm 0.3cm 0.3cm, clip=true]{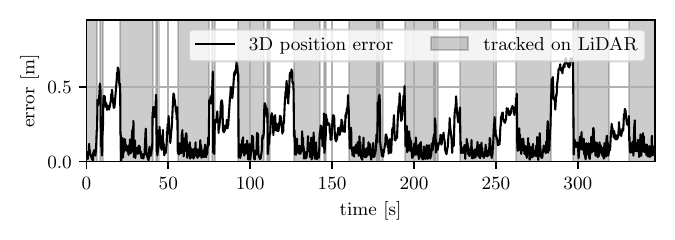}
  \caption{Relative localization error from \ac{NLOS} simulation.}
  \label{fig:sim_nlos_error}
  % \vspace{-1.0em}
\end{figure}

Fig.~\ref{fig:sim_nlos_error} shows the progression of the localization error of the proposed fusion approach with respect to the ground truth.
When the visibility was lost, the localization started drifting but was corrected each time when the visibility was re-established.
The position \ac{RMSE} was \SI{0.11}{m} when the \ac{UAV} was tracked by the primary \ac{UAV} (calculated from the localization data obtained at the time periods when the secondary \ac{UAV} was detected in the \ac{lidar} data) and \SI{0.35}{m} when it was out of sight.

\begin{figure}[t]
\centering
  \includegraphics[width=1.0\linewidth, trim=0.0cm 0cm 0cm 0cm, clip=true]{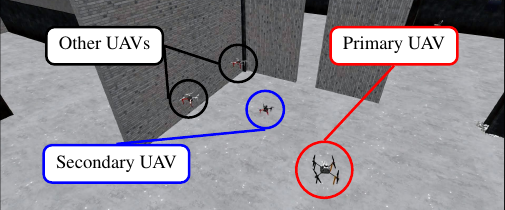}
  \caption{\ac{NLOS} experiment in the Gazebo simulator.}
\label{fig:sim_snapshot}
  % \vspace{-1.0em}
\end{figure}

\subsection{Real-world localization and guidance evaluation}\label{sec:realworld}
\label{sec:exp_guidance}

    To the best of the authors' knowledge, there are not yet any standard datasets containing multiple \acp{UAV} with \ac{lidar} data and heterogeneous localization methods, usable for testing such an approach, as proposed in this paper.
    Therefore, the raw data from the real-world flights performed in this work have been released online\footnote{\url{https://github.com/ctu-mrs/coop_uav_dataset}} to enable reproducibility of our results and provide data useful for developing such relative localization approaches to the research community.
    As there are no existing, openly-available algorithms utilizing the same hardware setup and alternative relative localization approaches require fundamentally different hardware on either one or both \acp{UAV}, making direct experimental comparison infeasible, the following quantitative evaluation of the proposed approach focuses mainly on the localization improvements provided by the fusion method with respect to the \ac{VIO}-only output.

\subsubsection{Experimental setup}\label{sec:guiding_setup}

\begin{figure}[t]
  \centering
  \includegraphics[width=1.0\linewidth, trim=0.0cm 0cm 0cm 0cm, clip=true]{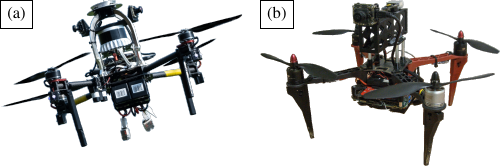}
  % \begin{tikzpicture}
  %   \node[anchor=north west,inner sep=0] (a) at (0, 0)
  %   {
  %     \includegraphics[width=0.48\linewidth, trim=0cm 0cm 0cm 0cm, clip=true]{figures/x500_smaller.png}
  %   };
  %   \node[fill=white,draw=black,text=black, anchor=north west] at (a.north west) {\footnotesize (a)};

  %   \node[anchor=north west,inner sep=0] (b) at (4.4cm, 0cm)
  %   {
  %     \includegraphics[width=0.48\linewidth, trim=0.0cm 0.0cm 0.0cm 0.0cm, clip=true]{figures/vio_drona.png}
  %   };
  %   \node[fill=white,draw=black,text=black, anchor=north west] at (b.north west) {\footnotesize (b)};

  % \end{tikzpicture}
  \caption{\ac{UAV} platforms employed in the real-world experiments: (a) primary \ac{UAV} carrying the Ouster 3D \ac{lidar}, (b) secondary \ac{UAV} with the monocular camera.%
  }
  \label{fig:drona}
  \vspace{-1.0em}
\end{figure}

The \ac{UAV} platforms are shown in Fig.~\ref{fig:drona}.
For a detailed description of the \ac{UAV} hardware, see~\cite{hertMRSDroneModular2023}.
The primary \ac{UAV} is built upon the X500 frame and carries the Intel NUC 10i7FNH onboard computer with the Intel Core i7 10710U CPU, 16 GB of RAM, and a Wi-Fi module.
The secondary \ac{UAV} is built upon the F330 frame and carries the Intel NUC 8i7BEH with the Intel Core i7-8559U CPU and 16 GB RAM.
Both \acp{UAV} carry the Pixhawk 4 \ac{FCU} containing the embedded attitude controller.

The primary \ac{UAV} carries the Ouster OS0-128 Rev~C 3D \ac{lidar}.
The \ac{lidar} weighs \SI{430}{g} without its top radial cap, has a 360$^\circ$ horizontal and 90$^\circ$ vertical \ac{FOV}, and produces scans with the resolution of $1024\times128$ beams at a rate of \SI{10}{Hz}.
The primary \ac{UAV} utilizes the LOAM SLAM algorithm~\cite{zhangLOAMLidarOdometry2014} for its self-localization.

The secondary \ac{UAV} carries the front-facing Bluefox MLC200Wc camera with the DSL217 fisheye lens.
The camera is rigidly connected to the ICM-42688-P \ac{IMU}.
The camera produces images at the rate of \SI{30}{Hz} at $752\times480$ resolution.
For self-localization, the secondary \ac{UAV} utilizes the VINS-Mono algorithm~\cite{qinVINSMonoRobustVersatile2018} with disabled loop closure detection.
The VINS-Mono algorithm processes the images at the rate of \SI{10}{Hz} and the \ac{IMU} at the rate of \SI{1000}{Hz}.
The \ac{UAV} is equipped with the Emlid Reach \ac{RTK} module for ground truth recording.

The software on board the \acp{UAV} is based on Ubuntu 20.04, \ac{ROS} 1, and the MRS UAV system~\cite{bacaMRSUAVSystem2021}.
Nimbro network is used for transporting the \ac{ROS} topics over the wireless network.
The system time of the onboard computers of the \acp{UAV} is synchronized using chrony~\footnote{\url{https://chrony.tuxfamily.org}} with the primary \ac{UAV} acting as a server and the secondary \ac{UAV} acting as a client.
All algorithms ran entirely on board the \acp{UAV}.
The \ac{RTK} data were collected for offline ground truth comparison and were not utilized in the control loop of the \acp{UAV} in any way.

The ground truth comparison was performed based on the trajectory evaluation methods described in~\cite{zhangTutorialQuantitativeTrajectory2018}.
Each evaluated trajectory was aligned with the \ac{RTK} ground truth data based on the first 20 seconds of each data series.
The error at each data point was calculated as the Euclidean distance between the aligned trajectory and the ground truth.
For each trajectory, the 2D and 3D \acp{ATE} were calculated as the total \ac{RMSE} between the ground truth and the aligned trajectory.

\subsubsection{Results}

\begin{figure}[t]
\centering
  \includegraphics[width=1.0\linewidth, trim=0.0cm 0cm 0cm 0cm, clip=true]{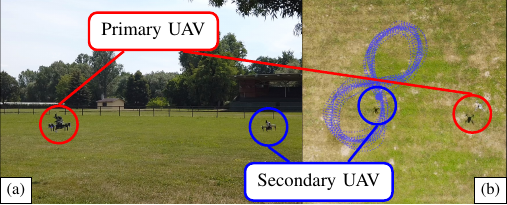}
  \caption{Side view (a) and top-down view (b) of the experimental verification. Ground-truth trajectory of the secondary \ac{UAV} is drawn over the image.}
\label{fig:real_exp}
  % \vspace{-1.0em}
\end{figure}

The secondary \ac{UAV} trajectory was predefined in the \ac{lidar} \ac{SLAM} frame, and the primary \ac{UAV} guided the secondary \ac{UAV} to follow the trajectory based on its commands.
Two trajectories were evaluated: a circle with \SI{0.5}{\meter\per\second} velocity and a figure eight with \SI{1}{\meter\per\second} velocity.
The \ac{UAV} faced the direction of its movement.
The primary \ac{UAV} was commanded by an operator to keep changing its pose while being stabilized based on its \ac{lidar} localization.
In the circular flight, the mutual \ac{UAV} distance ranged between \SIrange{3.2}{10.9}{m} with a mean of \SI{7.0}{m}.
In the figure-eight flight, the distance ranged between \SIrange{2.5}{9.8}{m} with a mean of \SI{6.3}{m}.
The algorithms proved capable of real-time usage on board the \acp{UAV} with the \ac{lidar}-based detector producing detections at the rate of \SI{10}{Hz} and the fusion algorithm taking less than \SI{5}{ms} to fuse them to produce each estimate.

% \begin{figure}[ht]
%   \centering
%   \includegraphics[width=0.9\linewidth, trim=0.0cm 0.46cm 0cm 1.11cm, clip=true]{figures/exp_circle_3d.pdf}
%   \caption{Trajectories from the circular flight.}
%   \label{fig:exp_circle_3d}
%   \vspace{-1.0em}
% \end{figure}

% \begin{figure}[ht]
%   \centering
%   \includegraphics[width=0.9\linewidth, trim=0.0cm 0.25cm 0cm 1.11cm, clip=true]{figures/exp_eight_3d.pdf}
%   \caption{Trajectories from the figure-eight flight.}
%   \label{fig:exp_eight_3d}
%   \vspace{-1.0em}
% \end{figure}

\begin{figure}[t]
  \centering
  \includegraphics[width=1.0\linewidth, trim=1.1cm 0.1cm 1.35cm 1.1cm, clip=true]{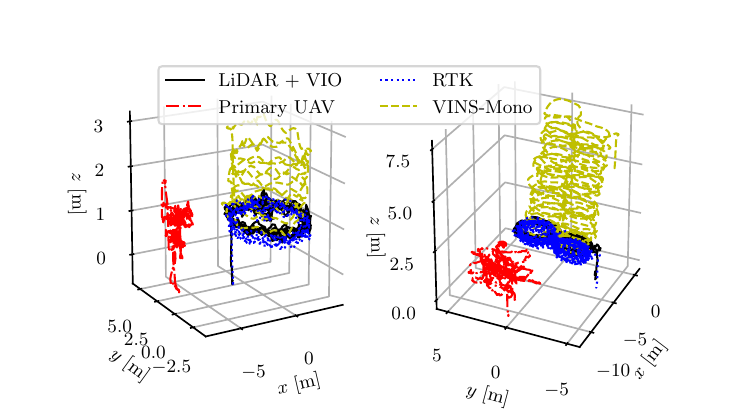}
  \caption{Trajectories from the real-world flights.}
  \label{fig:exp_combined_3d}
  % \vspace{-1.0em}
\end{figure}

\begin{figure}[t]
  \centering
  \includegraphics[width=1.0\linewidth, trim=0cm 0cm 0cm 0cm, clip=true]{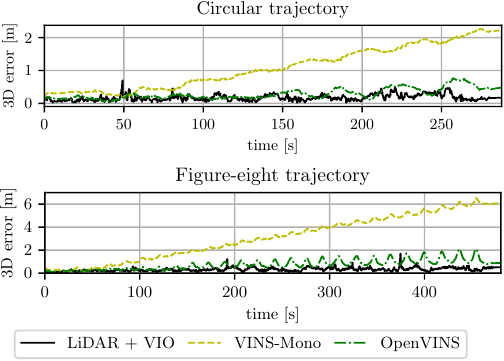}
  % \begin{tikzpicture}
  %   \node[anchor=north west,inner sep=0] (a) at (0cm, 2.8cm)
  %   {
  %     \includegraphics[width=1.0\linewidth, trim=0.3cm 0.3cm 0.3cm 0.3cm, clip=true]{figures/exp_circle_error.pdf}
  %   };
  %   % \node[fill=white,draw=black,text=black, anchor=north west] at (a.north west) {\footnotesize (a)};

  %   \node[anchor=north west,inner sep=0] (b) at (0.0cm, 0.0cm)
  %   {
  %     \includegraphics[width=1.0\linewidth, trim=0.25cm 0.2cm 0.2cm 0.2cm, clip=true]{figures/exp_eight_error.pdf}
  %   };
  %   % \node[fill=white,draw=black,text=black, anchor=north west] at (b.north west) {\footnotesize (b)};

  % \end{tikzpicture}
  \caption{Progression of 3D localization error over time from both of the performed trajectories.
  The error was calculated as the Euclidean distance between the \ac{RTK} ground truth and the respective aligned trajectory.}
  \label{fig:errors}
  % \vspace{-1.0em}
\end{figure}

The data from flights, aligned to a common reference frame, are shown in Fig.~\ref{fig:exp_combined_3d}.
In both cases, the \ac{VIO} output exhibited significant drift, mainly in the direction of the $z-$axis.
Despite the drift, the proposed method accurately tracked the secondary \ac{UAV} and enabled correct guidance of the \ac{UAV} along the desired trajectories.
Fig.~\ref{fig:errors} compares the localization error of the \ac{lidar}~+~\ac{VIO} fusion, VINS-Mono, and OpenVINS.
OpenVINS~\cite{Geneva2020ICRA} was run offline using the collected data for error comparison.
While the output of the \ac{VIO} methods was constantly drifting, the error of the \ac{lidar}~+~\ac{VIO} fusion was unaffected by this drift.
  The data shows that drift between two \ac{VIO} methods can significantly differ, even if they use the same data, making the drift values difficult to estimate beforehand.
  Furthermore, the amount of drift can change with texture in the scene, illumination, propeller-induced vibrations picked up by the \ac{IMU}, etc.

Table~\ref{tab:rmse} contains the \acp{ATE} of the \ac{lidar}~+~\ac{VIO} fusion and the tested \ac{VIO} methods.
Both the 3D error and the 2D error in the $xy$-plane are provided, as the VINS-Mono drift was most significant in the $z$-axis, and the \ac{lidar} vertical resolution is lower than its horizontal resolution.
The 3D \ac{ATE} of the \ac{lidar}~+~\ac{VIO} fusion was \SI{0.19}{\meter} in the circular flight and \SI{0.36}{\meter} in the figure-eight flight, significantly lower than the errors of the \ac{VIO} output.
The fusion error in the figure-eight flight was larger than that of the circular flight due to the greater velocity of the secondary \ac{UAV}.
The experimental evaluation showed that guidance based on the \ac{lidar}~+~\ac{VIO} fusion is capable of mitigating the effects of the \ac{VIO} drift on the flight of the secondary \ac{UAV}.

\begin{table}[h]
% \small
% \scriptsize
  \footnotesize
% \tiny
% \begin{center}
  \caption{\rev{Overall \acp{ATE} in meters, calculated as the \ac{RMSE} of all aligned data points.}}
  \begin{tabularx}{1.0\linewidth}{c *{8}{Y}} 
\toprule 
    \multirow{2}{*}{Trajectory} & \multicolumn{2}{c}{LiDAR + VIO} & \multicolumn{2}{c}{VINS-Mono} & \multicolumn{2}{c}{OpenVINS}\\ 
    \cmidrule(lr){2-3}  \cmidrule(lr){4-5} \cmidrule(lr){6-7}
     & 2D & 3D & 2D & 3D & 2D & 3D \\ \midrule
    \texttt{circle} & \textbf{0.13} & \textbf{0.19} & 0.49 & 1.21 & 0.28 & 0.32 \\ 
    \texttt{eight} & \textbf{0.22} & \textbf{0.36} & 1.44 & 3.55 & 0.68 & 0.83 \\ \midrule
    \rev{Average} & \rev{\textbf{0.18}} & \rev{\textbf{0.28}} & \rev{0.97} & \rev{2.38} & \rev{0.48} & \rev{0.58} \\ \bottomrule
\end{tabularx}
  \label{tab:rmse}
% \end{center}
  % \vspace{-1.5em}
\end{table}

\subsection{Real-world cooperative mapping scenario}
\label{sec:exp_coop_warehouse}

\begin{figure}[t]
  \centering
  \includegraphics[width=1.0\linewidth, trim=0.0cm 0.0cm 0cm 1.0cm, clip=true]{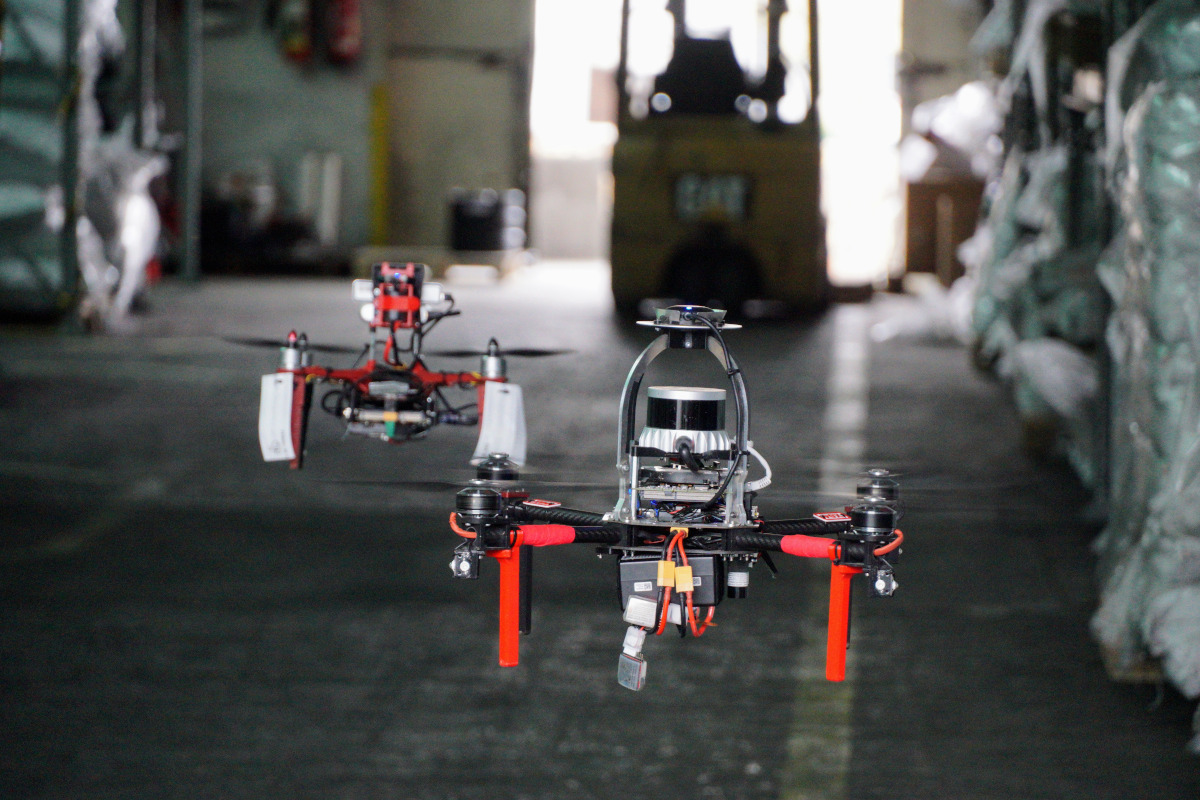}
  \caption{The \acp{UAV} mapping an industrial warehouse.}
  \label{fig:rudna_drony}
  % \vspace{-1.0em}
\end{figure}

\begin{figure*}[t]
\centering
  \includegraphics[width=1.0\linewidth, trim=0.0cm 0cm 0cm 0cm, clip=true]{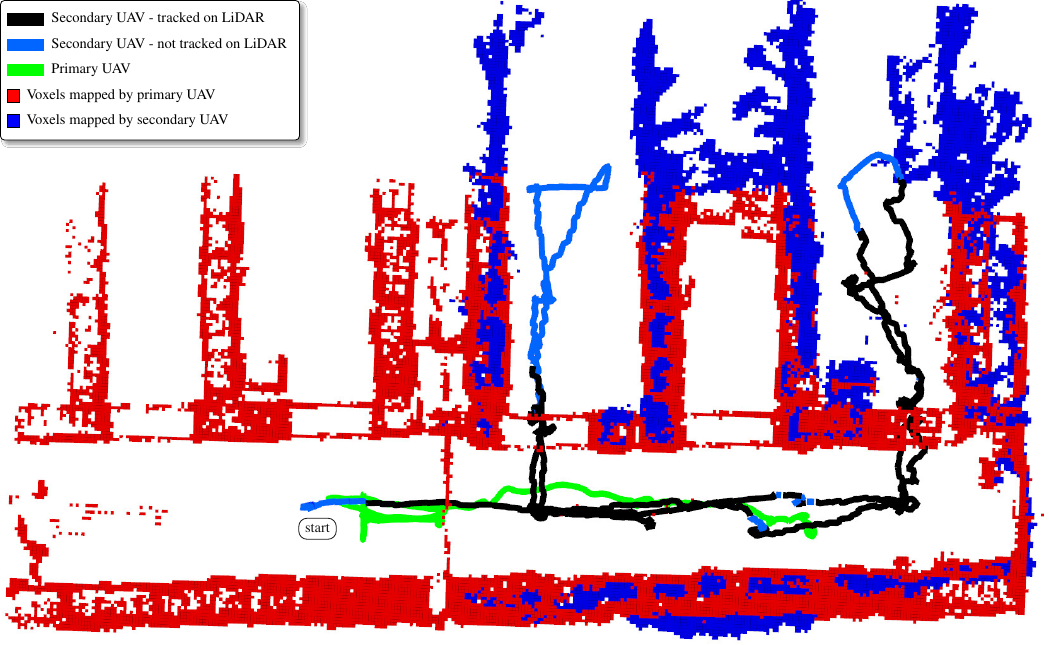}
  \caption{\ac{UAV} trajectories and projection of the global occupancy map obtained during the real-world cooperative mapping experiment in an industrial warehouse.}
\label{fig:rudna_map}
  % \vspace{-1.0em}
\end{figure*}

The proposed relative localization method was applied to a cooperative mapping scenario to enable occupancy mapping of an industrial warehouse using a heterogeneous \ac{UAV} team (see Fig.~\ref{fig:rudna_drony} and the multimedia attachment\footnote{\url{https://mrs.fel.cvut.cz/coop-fusion}}).

The experiment demonstrated that the proposed method is sufficiently accurate to enable precise cooperative mapping of a complex real-world environment.
The deployed \acp{UAV} were identical to the platforms described in Sec.~\ref{sec:guiding_setup}, except that the secondary \ac{UAV} was equipped with RealSense T265 tracking camera for \ac{VIO} and RealSense D435 depth camera for dense occupancy mapping.
The secondary \ac{UAV} utilized OpenVINS~\cite{Geneva2020ICRA} as the \ac{VIO} algorithm.
\bl{The secondary \ac{UAV} was equipped with reflective markers on its legs to provide an additional safety feature for robust performance of the \ac{lidar}-based detection algorithm in the cluttered environment.
The size of the reflective markers was designed taking into account the parameters of the Ouster OS0-128 \ac{lidar}, such that the markers would be larger than the spread between the adjacent \ac{lidar} beams up to the distance of \SI{8}{m}.}
Both \acp{UAV} carried the Intel NUC 10i7FNH onboard computer.

In the experiment, the primary \ac{UAV} stayed in a wide main corridor, while the secondary \ac{UAV} was sent to map narrow side corridors.
Desired goals for both \acp{UAV} were selected by the operator during the experiment based on the online-generated map.
The primary \ac{UAV} flew to selected goal points and guided the secondary \ac{UAV} for brief periods before the secondary \ac{UAV} flew into the side corridors.
After the secondary \ac{UAV} flew into each side corridor, it performed autonomous exploration of the area for a predefined amount of time.
While exploring the side corridors, the secondary \ac{UAV} was not tracked on \ac{lidar} for part of its flight due to occlusions (see Fig.~\ref{fig:rudna_map}), therefore its estimated pose was updated only based on the incoming \ac{VIO} data.

Each \ac{UAV} constructed a local occupancy map online from its onboard sensors.
The secondary \ac{UAV} transmitted its local map to the primary \ac{UAV} over the wireless network at the rate of \SI{0.5}{Hz}.
The local maps were merged together onboard the primary \ac{UAV} using the output of the proposed relative localization method.
The obtained global occupancy map is shown in Fig.~\ref{fig:rudna_map}.
The results of the experiment qualitatively evaluate the accuracy of the proposed method and showcase that it can run on board the \acp{UAV} as a part of a complex software stack aimed for, e.g., autonomous cooperative exploration and mapping.

\subsection{Tuning methodology, parameter sensitivity, and measurement delay influence analysis}
\begin{figure}[t]
  \centering
  \includegraphics[width=1.0\linewidth, trim=0cm 0cm 0cm 0cm, clip=true]{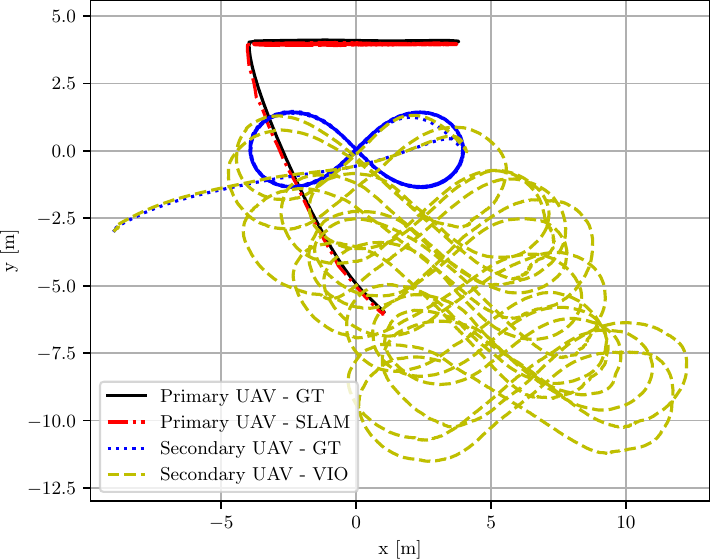}
  \caption{Ground-truth and estimated trajectories from the tuning and parameter sensitivity analysis.}
  \label{fig:tuning_trajectories}
  % \vspace{-1.0em}
\end{figure}

To tune the parameters of the proposed method and evaluate its robustness, a simulation was designed in the Gazebo robotic simulator.
The secondary \ac{UAV} was following a figure-eight trajectory, while the primary \ac{UAV} was following a line trajectory next to it (see Fig.~\ref{fig:tuning_trajectories}).
The LOAM algorithm was run on the \ac{lidar} data of the primary \ac{UAV} and the OpenVINS algorithm was run on the monocular camera data of the secondary \ac{UAV} for localization.
While the \ac{lidar}-based localization was reliable throughout the flight, the \ac{VIO} exhibited significant drift.
Data recorded in this simulation were used to tune the algorithm and perform the parameter sensitivity and measurement delay influence analyses.

% - vyplotit gt trajektorie + lidar a vio odometrie

\subsubsection{NLS-based estimator}

\begin{figure}[t]
  \centering
  \includegraphics[width=1.0\linewidth, trim=0cm 0cm 0cm 0cm, clip=true]{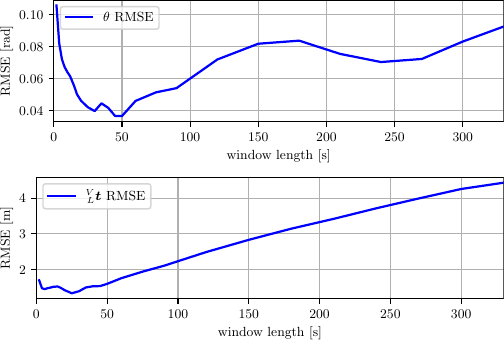}
  % \begin{tikzpicture}
  %   \node[anchor=north west,inner sep=0] (a) at (0cm, 3.0cm)
  %   {
  %     \includegraphics[width=1.0\linewidth, trim=0.0cm 0.0cm 0.0cm 0.0cm, clip=true]{figures/window_length_theta.pdf}
  %   };
  %   % \node[fill=white,draw=black,text=black, anchor=north west] at (a.north west) {\footnotesize (a)};

  %   \node[anchor=north west,inner sep=0] (b) at (0.0cm, 0.0cm)
  %   {
  %     \includegraphics[width=1.0\linewidth, trim=0.0cm 0.0cm 0.0cm 0.0cm, clip=true]{figures/window_length_trans.pdf}
  %   };
  %   % \node[fill=white,draw=black,text=black, anchor=north west] at (b.north west) {\footnotesize (b)};

  % \end{tikzpicture}
  \caption{Dependency of the relative transformation $\frames{V}{L}{\mat{T}}$ RMSE on the sliding-window length.}
  \label{fig:window_length}
  % \vspace{-1.0em}
\end{figure}

Fig.~\ref{fig:window_length} shows the dependency of the relative transformation $\frames{V}{L}{\mat{T}}$ \ac{RMSE} on the length of the sliding window in the \ac{NLS}-based estimator.
The \ac{RMSE} was measured for 26 length values ranging from \SI{2} to \SI{330}{s}.
For short sliding windows, the \ac{RMSE} was large as the estimate was corrupted by measurement noise.
The \ac{RMSE} reached the lowest values for \SI{25}{s} in the case of the translational error and \SI{50}{s} in the case of the rotational error.
For longer sliding windows, the \ac{RMSE} gradually increased due to increasing lag of the estimate.
Based on the results, setting the sliding window length to a value in the range between \SI{20}{s} and \SI{50}{s} is advisable to obtain reliable estimator performance.

\subsubsection{LKF-based tracker}

The \ac{LKF}-based tracker was tuned by analysing the performance of the algorithm with respect to the ground truth data in the simulated scenario.
The measurement noise parameters, namely the detection variance $\sigma^2_d$, relative heading variance $\sigma^2_\theta$, VIO relative vector variance $\sigma^2_r$, VIO linear velocity variance $\sigma^2_V$, VIO heading variance $\sigma^2_\phi$, and VIO heading rate variance $\sigma^2_\omega$, were measured in the simulation and set accordingly.
The process noise parameters include the linear velocity variance $\sigma^2_v$ and the angular velocity variance $\sigma^2_\omega$.
The process noise parameters were tuned by analysing the \ac{ANEES}~\cite{barshalomestimation} of the final estimate with respect to the ground-truth data and tuning the parameters to achieve consistent estimation performance, i.e., so that the \ac{ANEES} over the dataset corresponds to the degrees of freedom of the state vector.
% Finally, the Mahalanobis distance threshold was set to \textbf{TODO} to achieve reliable outlier rejection while fusing all the valid measurements.

\begin{figure}[t]
  \centering
  \includegraphics[width=1.0\linewidth, trim=0cm 0cm 0cm 0cm, clip=true]{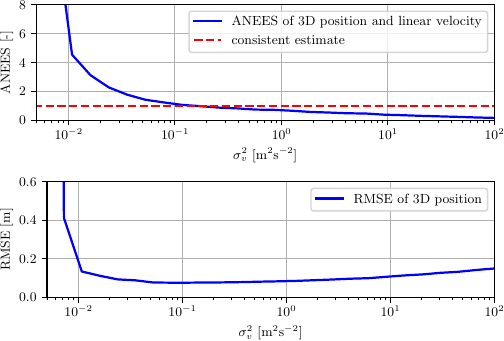}
  % \begin{tikzpicture}
  %   \node[anchor=north west,inner sep=0] (a) at (0cm, 3.0cm)
  %   {
  %     \includegraphics[width=1.0\linewidth, trim=0.0cm 0.0cm 0.0cm 0.0cm, clip=true]{figures/q_tuning_nees_pos.pdf}
  %   };
  %   % \node[fill=white,draw=black,text=black, anchor=north west] at (a.north west) {\footnotesize (a)};

  %   \node[anchor=north west,inner sep=0] (b) at (0.0cm, 0.0cm)
  %   {
  %     \includegraphics[width=1.0\linewidth, trim=0.0cm 0.0cm 0.0cm 0.0cm, clip=true]{figures/q_tuning_rmse_pos.pdf}
  %   };
  %   % \node[fill=white,draw=black,text=black, anchor=north west] at (b.north west) {\footnotesize (b)};

  % \end{tikzpicture}
  \caption{Dependencies of the positional \ac{RMSE} and of the \ac{ANEES} of 3D position and linear velocity on $\sigma_v^2$.
  The \ac{ANEES} was normalized by the number of \acp{DOF}.
  \ac{ANEES} of one corresponds to a consistent estimate.}
  \label{fig:q_tuning_pos}
  % \vspace{-1.0em}
\end{figure}

\begin{figure}[t]
  \centering
  \includegraphics[width=1.0\linewidth, trim=0cm 0cm 0cm 0cm, clip=true]{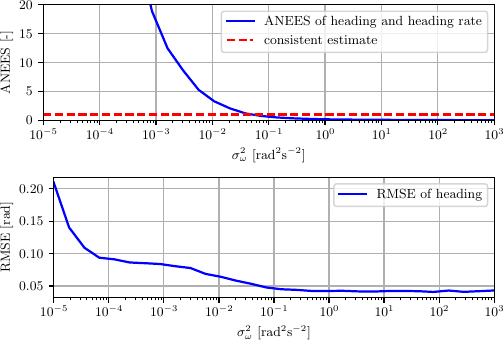}
  % \begin{tikzpicture}
  %   \node[anchor=north west,inner sep=0] (a) at (0cm, 3.0cm)
  %   {
  %     \includegraphics[width=1.0\linewidth, trim=0.0cm 0.0cm 0.0cm 0.0cm, clip=true]{figures/q_tuning_nees_yaw.pdf}
  %   };
  %   % \node[fill=white,draw=black,text=black, anchor=north west] at (a.north west) {\footnotesize (a)};

  %   \node[anchor=north west,inner sep=0] (b) at (0.0cm, 0.0cm)
  %   {
  %     \includegraphics[width=1.0\linewidth, trim=0.0cm 0.0cm 0.0cm 0.0cm, clip=true]{figures/q_tuning_rmse_yaw.pdf}
  %   };
  %   % \node[fill=white,draw=black,text=black, anchor=north west] at (b.north west) {\footnotesize (b)};

  % \end{tikzpicture}
  \caption{Dependencies of the heading \ac{RMSE} and of the \ac{ANEES} of the heading and heading rate on $\sigma_\omega^2$.
  The \ac{ANEES} was normalized by the number of \acp{DOF}.
  \ac{ANEES} of one corresponds to a consistent estimate.}
  \label{fig:q_tuning_yaw}
  % \vspace{-1.0em}
\end{figure}

To tune the parameter $\sigma_v^2$ and to determine the effect of the parameter setting on the tracker output, \ac{UAV} pose estimator was run on the simulation dataset for a set of 30 logarithmically-spaced values of $\sigma_v^2$ ranging from \SI{0.01} to \SI{100}{m^2s^{-2}}.
Fig.~\ref{fig:q_tuning_pos} shows the dependency of the positional \ac{RMSE} and of the \ac{ANEES} of 3D position and linear velocity on different values of $\sigma_v^2$.
The best performance was obtained for $\sigma_v^2$ close to \SI{0.1}{m^2s^{-2}}.
For lower values of $\sigma_v^2$, the tracker exhibits increasing lag of the estimate, with too large lag resulting in loss of tracking as the distance of the incoming detections from the estimate becomes larger than the association threshold.
For larger values of $\sigma_v^2$, the resulting estimate is underconfident and the measurement noise is not filtered properly.
Fig.~\ref{fig:q_tuning_yaw} shows the dependency of the heading \ac{RMSE} and \ac{ANEES} on the values of $\sigma^2_\omega$, ranging between $10^{-5}$ and $10^3$~$\mathrm{rad}^2\mathrm{s}^{-2}$.
The best \ac{ANEES} was achieved for the value of \SI{0.05}{rad^2s^{-2}}.
The heading measurements in the simulation were not subject to significant Gaussian noise, therefore the \ac{RMSE} exhibited low values even for higher values of $\sigma^2_\omega$.

\subsubsection{Measurement delay influence analysis}

\begin{figure}[t]
  \centering
  \includegraphics[width=1.0\linewidth, trim=0cm 0cm 0cm 0cm, clip=true]{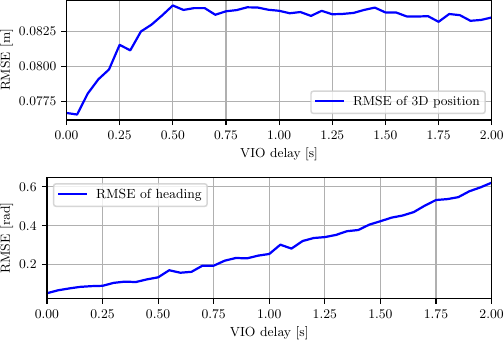}
  % \begin{tikzpicture}
  %   \node[anchor=north west,inner sep=0] (a) at (0cm, 3.0cm)
  %   {
  %     \includegraphics[width=1.0\linewidth, trim=0.0cm 0.0cm 0.0cm 0.0cm, clip=true]{figures/robustness_vio_delay0.pdf}
  %   };
  %   % \node[fill=white,draw=black,text=black, anchor=north west] at (a.north west) {\footnotesize (a)};

  %   \node[anchor=north west,inner sep=0] (b) at (0.0cm, 0.0cm)
  %   {
  %     \includegraphics[width=1.0\linewidth, trim=0.0cm 0.0cm 0.0cm 0.0cm, clip=true]{figures/robustness_vio_delay1.pdf}
  %   };
  %   % \node[fill=white,draw=black,text=black, anchor=north west] at (b.north west) {\footnotesize (b)};

  % \end{tikzpicture}
  \caption{Dependency of positional and heading \ac{RMSE} on the delay of the incoming \ac{VIO} measurements.}
  \label{fig:robustness_vio_delay}
  % \vspace{-1.0em}
\end{figure}

\begin{figure}[t]
  \centering
  \includegraphics[width=1.0\linewidth, trim=0cm 0cm 0cm 0cm, clip=true]{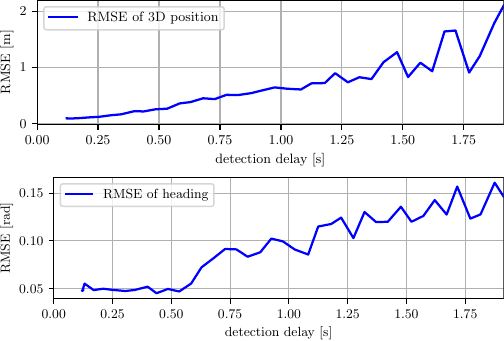}
  % \begin{tikzpicture}
  %   \node[anchor=north west,inner sep=0] (a) at (0cm, 3.0cm)
  %   {
  %     \includegraphics[width=1.0\linewidth, trim=0.0cm 0.0cm 0.0cm 0.0cm, clip=true]{figures/robustness_lidar_delay0.pdf}
  %   };
  %   % \node[fill=white,draw=black,text=black, anchor=north west] at (a.north west) {\footnotesize (a)};

  %   \node[anchor=north west,inner sep=0] (b) at (0.0cm, 0.0cm)
  %   {
  %     \includegraphics[width=1.0\linewidth, trim=0.0cm 0.0cm 0.0cm 0.0cm, clip=true]{figures/robustness_lidar_delay1.pdf}
  %   };
  %   % \node[fill=white,draw=black,text=black, anchor=north west] at (b.north west) {\footnotesize (b)};

  % \end{tikzpicture}
  \caption{Dependency of positional and heading \ac{RMSE} on the delay of the \ac{lidar} detections.}
  \label{fig:robustness_lidar_delay}
  % \vspace{-1.0em}
\end{figure}

Fig.~\ref{fig:robustness_vio_delay} analyzes the dependency of localization error on the delay of the incoming \ac{VIO} measurements.
The simulation was performed for 41 communication-delay values ranging from \SI{0} to \SI{2}{s}.
The positional accuracy of the proposed method depends mainly on the accuracy of the \ac{lidar} detections with the \ac{VIO} measurements improving the performance when the \ac{lidar} detections are unavailable.
For the \ac{VIO} delay values from \SI{0} to \SI{0.5}{s}, the positional \ac{RMSE} was slowly rising as the amount of \ac{VIO} measurements usable in the case of \ac{lidar} detection dropouts or large detection delay was decreasing.
In the simulations, there were no significant \ac{lidar} detection dropouts longer than \SI{0.5}{s}, therefore the positional \ac{RMSE} remained approximately constant for larger \ac{VIO} delay values.
The heading of the \ac{UAV} is observed solely from the \ac{VIO} measurements.
Therefore, the heading accuracy kept growing with the increasing \ac{VIO} delay.

Fig.~\ref{fig:robustness_lidar_delay} shows the dependency of the localization error on the average delay of the \ac{lidar} detections.
The \ac{lidar} detection delay influences mainly the positional error, which grew linearly from \SI{0.09}{m} for the average delay of \SI{0.12}{s} to \SI{2.10}{m} for the average delay of \SI{1.91}{s}.
The heading error slightly increased with growing detection delay due to the growing inaccuracy in the estimated relative transformation $\frames{V}{L}{\mat{T}}$.

\begin{figure}
\centering
  \includegraphics[width=1.0\linewidth, trim=0.0cm 0.0cm 0.0cm 0.0cm, clip=true]{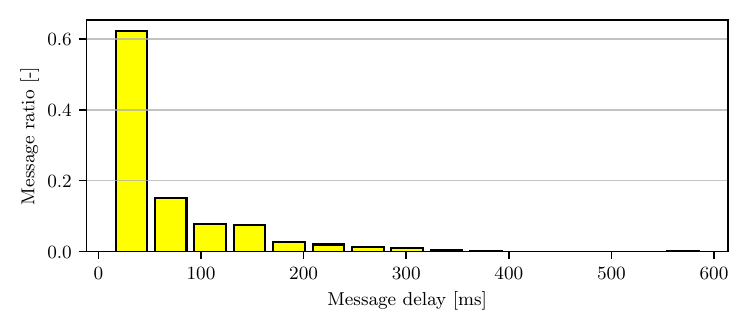}
  \caption{\bl{Histogram of \ac{VIO} message delays from the cooperative mapping experiment.}}
\label{fig:message_delay}
  % \vspace{-1.5em}
\end{figure}

\bl{
\subsection{Discussion}
\subsubsection{Scaling to more secondary UAVs}
This work focused on the fundamental design of the fusion and guidance method for a single primary and secondary \ac{UAV} pair and an exhaustive experimental study on scaling the approach to larger swarms is reserved for future work.
Nevertheless, we shall discuss two important aspects for scaling the approach to multiple \acp{UAV}: the communication requirements and the initialization of the approach.
Table~\ref{tab:table_communication} shows the communication requirements recorded in the real-world experiments in Sec.~\ref{sec:exp_guidance} and Sec.~\ref{sec:exp_coop_warehouse}.
The communication from the secondary \ac{UAV} to the primary \ac{UAV} is constant, and the bandwidth is proportional to the rate of transmitted messages.
In the experiments, the bandwidth for a single \ac{UAV} was \SI{1.45}{KB\per s}.
The communication from the primary \ac{UAV} to the secondary \ac{UAV} depends on the rate of communication and on the number of points in the transmitted trajectory messages.
In the implementation, each trajectory point was represented by 32 bytes.
In the experiments, the bandwidth ranged from \SI{0.61}{KB\per s} to \SI{30.53}{KB\per s}.
The communication requirements will scale linearly with the number of secondary \acp{UAV}, since all the secondary \acp{UAV} will need to receive and transmit data to the primary \ac{UAV}, but the secondary \acp{UAV} will not communicate among themselves, and will scale linearly with the number of points in the reference trajectory.
The TP-Link TL-WR902AC WiFi router was utilized in the experiments.
The router has a maximum data transfer rate of \SI{433}{Mb\per s} over a \SI{5}{GHz} network.
Assuming the maximum bandwidth required by the approach to be \SI{32}{KB\per s} and no other data transmission required, the router is theoretically capable of supporting up to 1691 such transmissions.
As such, the communication requirements are well below the capabilities of current wireless communication modules and will not restrict scaling of the approach to multiple secondary \acp{UAV}.

For initialization of the approach, the proposed approach requires that each secondary UAV performs a separate initialization maneuver to initialize the method as described in Sec.~\ref{sec:initialization}.
The number of required initialization maneuvers scales linearly with the number of secondary \acp{UAV}.

As shown in~\cite{vrbaOnboardLiDARbasedFlying2023}, the employed \ac{UAV} detection approach is capable of detecting multiple targets simultaneously and does not impede scaling the approach to multiple secondary \acp{UAV}.

\subsubsection{Operational limits of the system}
As shown in Sec.~\ref{sec:sim_drift_compensation}, for the tested configuration the guidance of the secondary \ac{UAV} based on the \ac{lidar} detections was able to compensate for constant-velocity odometry drift of the secondary \ac{UAV} for drift velocities of up to \SI{0.7}{m\per s} and with one failure for the value of \SI{0.8}{m\per s}.
In case of a complete and sudden divergence of the \ac{VIO} estimate, resulting in larger drift, the proposed method would not be able to compensate for it.
The proposed approach is designed for drift compensation rather than failure recovery.
Consequently, handling a catastrophic \ac{VIO} failure, e.g., a total loss of tracking, is considered outside the current scope of this work.
The method also fully relies on the yaw estimate from the \ac{VIO}, as the \ac{lidar} detections do not provide any information about the orientation of the secondary \ac{UAV}.
Therefore, in case of a complete failure of the \ac{VIO}, the fusion method would not be able to obtain the yaw orientation of the secondary \ac{UAV}.
As this paper focuses on the fundamental estimation framework, a fallback solution for providing the secondary \ac{UAV}'s yaw is outside the scope of this work.
In a practical application in a production-ready system, such a problem might be solved, e.g., by fusion of magnetometer data.

The maximum range of the method depends on the maximum range of the LiDAR sensor.
In case of the utilized Ouster OS0-128 Rev C, the maximum range was 45 meters at 80\% Lambertian reflectivity with >90\% detection probability\footnote{\url{https://data.ouster.io/downloads/datasheets/datasheet-revc-v2p5-os0.pdf}, accessed 23/07/2025}.
Also, with increasing distances, the secondary UAV may not be detected if it is smaller than the spread between two adjacent LiDAR beams and falls between that spread.
Such details regarding the LiDAR-based UAV detection are described in~\cite{vrbaOnboardLiDARbasedFlying2023}.

Large communication latency can be compensated for by the recalculating history buffer~\cite{pritzlAdaptiveAltitudeEstimation} used by the approach.
Delayed measurements will be fused correctly, assuming that they are not older than the size of the history buffer.
To showcase the real-world communication delays the system faced in the experiments, Fig.~\ref{fig:message_delay} shows the \ac{VIO} message delays from the real-world deployment in the cooperative mapping scenario described in Sec.~\ref{sec:exp_coop_warehouse}.
The \ac{VIO} message delay ranged from \SI{13}{ms} to \SI{589}{ms} with the median delay of \SI{39}{ms}.
To account for packet loss, the estimator utilizes the constant-velocity model, which predicts the state of the secondary \ac{UAV} if no measurements are received.

Under poor visibility, the method would benefit from the robustness of the \ac{lidar} sensor, which can work even in darkness.
However, the method would be unable to compensate for total \ac{VIO} failure, as explained above.
In case of occlusions and \ac{NLOS} situations, the method will keep estimating the pose of the secondary \ac{UAV} based on the incoming \ac{VIO} data, as demonstrated in Sec.~\ref{sec:sim_nlos}.
When the line-of-sight gets re-established, new detections will be associated to the estimate, and \ac{VIO} drift will be corrected.
In case of too large \ac{VIO} drift during the \ac{NLOS} situation, the estimate will be re-initialized after a sufficient amount of detections is obtained.
The re-initialization is retried periodically while the estimator is receiving \ac{VIO} data over the wireless network.

}

\begin{table}[t]
% \small
\scriptsize
\begin{center}
  \caption{\bl{Analysis of the wireless communication requirements during the real-world experiments.}}
\begin{tabularx}{0.95\linewidth}{l *{4}{Y}} 
\toprule 
  & Rate [Hz] & Message size [KB] (min/max) & Bandwidth [KB/s] (min/max) \\
\midrule
Sec.$\rightarrow$Prim. (odometry) & 2 & 0.725 / 0.725 & 1.45 / 1.45 \\
  Prim.$\rightarrow$Sec. - Exp.~\ref{sec:exp_guidance} & 5 & 0.121 / 6.105 & 0.61 / 30.53 \\
  Prim.$\rightarrow$Sec. - Exp.~\ref{sec:exp_coop_warehouse} & 5 & 0.132 / 2.457 & 0.66 / 12.29 \\
\bottomrule
\end{tabularx}
  \label{tab:table_communication}
\end{center}
  \vspace{-1.5em}
\end{table}

% trajectory - warehouse
% total_max_msg_size: 2.457 KB --> 12.29 KB / s
% total_min_msg_size: 0.185 KB

% total_max_msg_size: 0.228 KB
% total_min_msg_size: 0.132 KB --> 0.66 bw

%%}

%%{ CONCLUSIONS

\section{Conclusions}
    % \vspace{-0.3em}
\label{sec:conclusion}

An approach for cooperative guidance of a micro-scale \ac{UAV} relying on a novel technique of relative localization was proposed in this paper.
We proposed a heterogeneous \ac{UAV} team consisting of a \ac{lidar}-equipped primary \ac{UAV} and a small camera-equipped secondary \ac{UAV}, combining the benefits of both sensor types.
To achieve the desired tight cooperation of \acp{UAV} with different sensory modalities, we proposed a novel approach for fusing the \ac{lidar} relative localization with \ac{VIO} output on board the primary \ac{UAV} to obtain accurate pose of the secondary \ac{UAV}.
We demonstrated that the resulting pose estimate can be used to guide the secondary \ac{UAV} along desired trajectories defined in the reference frame of the primary \ac{UAV}.
The performance of the proposed approach was verified and quantitatively evaluated in simulations and real-world experiments.
The results showed the superior accuracy of the fusion output to the pure \ac{VIO} data, successfully demonstrated the capability to guide the secondary \ac{UAV} along the desired trajectories, and showcased the usability of the proposed approach in a cooperative mapping scenario.

%%}

\begin{acronym}
  % \acro{ML}[ML]{machine learning}
  \acro{GPS}[GPS]{Global Positioning System}
  \acro{CNN}[CNN]{Convolutional Neural Network}
  \acro{MAV}[MAV]{Micro Aerial Vehicle}
  \acro{UAV}[UAV]{Unmanned Aerial Vehicle}
  \acro{UGV}[UGV]{Unmanned Ground Vehicle}
  \acro{UV}[UV]{ultraviolet}
  \acro{UVDAR}[\emph{UVDAR}]{UltraViolet Direction And Ranging}
  \acro{UT}[UT]{Unscented Transform}
  \acro{GNSS}[GNSS]{Global Navigation Satellite System}
  \acro{RTK}[RTK]{Real-time kinematic}
  \acro{MOCAP}[mo-cap]{Motion capture}
  \acro{ROS}[ROS]{Robot Operating System}
  \acro{MPC}[MPC]{Model Predictive Control}
  \acro{MBZIRC}[MBZIRC 2020]{Mohamed Bin Zayed International Robotics Challenge 2020}
  \acro{MBZIRC19}[MBZIRC 2019]{Mohamed Bin Zayed International Robotics Challenge 2019}
  \acro{FOV}[FOV]{Field Of View}
  \acrodefplural{FOV}[FOVs]{Fields of View}
  \acro{ICP}[ICP]{Iterative closest point}
  \acro{FSM}[FSM]{Failure recovery and Synchronization jobs Manager}
  \acro{IMU}[IMU]{Inertial Measurement Unit}
  \acro{EKF}[EKF]{Extended Kalman Filter}
  \acro{LKF}[LKF]{Linear Kalman Filter}
  \acro{KF}[KF]{Kalman Filter}
  \acro{COTS}[COTS]{Commercially Available Off-the-Shelf}
  \acro{ESC}[ESC]{Electronic Speed Controller}
  \acro{lidar}[LiDAR]{Light Detection and Ranging}
  % \acrodefplural{lidar}[LiDARs]{Light Detection and Ranging}
  \acro{SLAM}[SLAM]{Simultaneous Localization and Mapping}
  \acro{SEF}[SEF]{Successive Edge Following}
  \acro{IEPF}[IEPF]{Iterative End-Point Fit}
  \acro{USAR}[USAR]{Urban Search and Rescue}
  \acro{SAR}[SAR]{Search and Rescue}
  \acro{ROI}[ROI]{Region of Interest}
  \acro{WEC}[WEC]{Window Edge Candidate}
  \acro{UAS}[UAS]{Unmanned Aerial System}
  \acro{VIO}[VIO]{Visual-Inertial Odometry}
  \acro{DOF}[DOF]{Degree of Freedom}
  \acrodefplural{DOF}[DOFs]{Degrees of Freedom}
  \acro{LTI}[LTI]{Linear Time-Invariant}
  \acro{FCU}[FCU]{Flight Control Unit}
  \acro{UWB}[UWB]{Ultra-wideband}
  \acro{ICP}[ICP]{Iterative Closest Point}
  \acro{NIS}[NIS]{Normalized Innovations Squared}
  \acro{LRF}[LRF]{Laser Rangefinder}
  \acro{RMSE}[RMSE]{Root Mean Squared Error}
  \acro{VINS}[VINS]{Vision-aided Inertial Navigation Systems}
  \acro{VSLAM}[VSLAM]{Visual Simultaneous Localization and Mapping}
  \acro{NLS}[NLS]{Non-linear Least Squares}
  \acro{NTP}[NTP]{Network Time Protocol}
  \acro{ATE}[ATE]{Absolute Trajectory Error}
  \acro{MSCKF}[MSCKF]{Multi-State Constraint Kalman Filter}
  \acro{NLOS}[NLOS]{Non-Line-of-Sight}
  \acro{SubT}[SubT]{Subterranean}
  \acro{DARPA}[DARPA]{Defense Advanced Research Projects Agency}
  \acro{KITTI}[KITTI]{Karlsruhe Institute of Technology and Toyota Technological Institute}
  \acro{NEES}[NEES]{Normalized Estimation Error Squared}
  \acro{ANEES}[ANEES]{Average Normalized Estimation Error Squared}
\end{acronym}

\bibliographystyle{IEEEtran}
% \bibliography{main}

% \bibliographystyle{elsarticle-num}
\bibliography{main}

\begin{IEEEbiography}[{\includegraphics[width=1in,height=1.25in,clip,keepaspectratio]{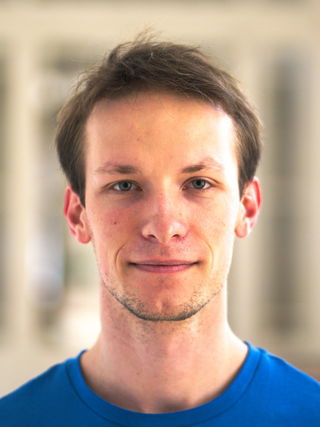}}]{V\'aclav Pritzl} received the M.Sc. degree in cybernetics and robotics from the Czech Technical University in Prague (CTU in Prague), Czech Republic, in 2020, where he is currently working toward the Ph.D. degree in cooperative multi-UAV navigation in GNSS-denied environments.
He is a member of the Multi-robot Systems Lab, CTU in Prague.
  He has authored or coauthored ten publications in conferences and impacted journals with >200 citations indexed by Scholar and h-index 8. His research interests include cooperative navigation of teams of UAVs in GNSS-denied environments. He was a member of CTU-UPENN-NYU team in the MBZIRC 2020.
\end{IEEEbiography}

\begin{IEEEbiography}[{\includegraphics[width=1in,height=1.25in,clip,keepaspectratio]{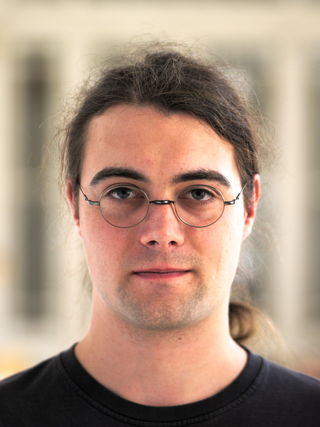}}]{Matou\v{s} Vrba} received the Ph.D. degree in detection and localization for autonomous aerial vehicles from the Czech Technical University in Prague (CTU in Prague), Czech Republic, in 2024.
Since 2018, he is a member of the Multi-robot Systems Lab, CTU Prague.
He has authored or coauthored 15 publications in conferences and impacted journals with >700 citations indexed by Scholar and h-index 13.
His research interests include machine perception methods for marker-less mutual localization of UAVs. 
He was a Member of the CTU-UPENN-NYU team in the MBZIRC 2020 and the CTU-CRAS-NORLAB team in the DARPA SubT competition.
\end{IEEEbiography}

\begin{IEEEbiography}[{\includegraphics[width=1in,height=1.25in,clip,keepaspectratio]{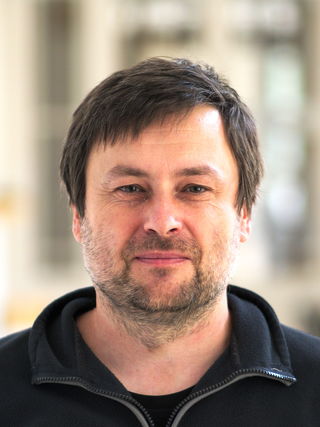}}]{Petr \v{S}t\v{e}p\'an} received the Ph.D. degree in sensor fusion for mapping from the Czech Technical University in Prague (CTU Prague), Prague, Czech Republic, in 2002.
He is currently with the Multi-Robot Systems lab, CTU Prague, where he focuses on sensor fusion, mapping, localization, and planning for unmanned aerial vehicles.
He has also been involved in industrial projects and the H2020 AerialCore project.
He is a coauthor of >40 publications in conferences and impacted journals with >700 citations indexed by Scholar and h-index 11. He was a member of CTU-UPenn-UoL and CTU-UPENN-NYU teams in the MBZIRC 2017 and MBZIRC 2020 robotic competitions in Abu Dhabi.
\end{IEEEbiography}

\begin{IEEEbiography}[{\includegraphics[width=1in,height=1.25in,clip,keepaspectratio]{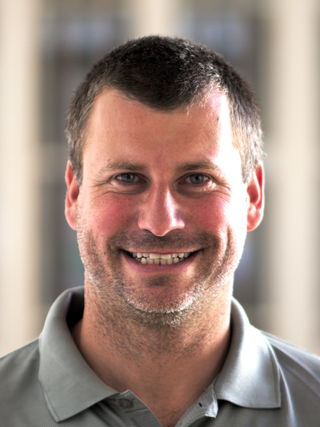}}]{Martin Saska} received the Ph.D. degree in trajectory planning and optimal control for formations of autonomous robots from the University of Wuerzburg, Wuerzburg, Germany, in 2010, within the Ph.D. program of Elite Network of Bavaria.
He founded and heads the Multi-robot Systems group at the Czech Technical University in Prague with more than 40 researchers.
He was a Visiting Scholar with the University of Illinois at Urbana-Champaign, Champaign, IL, USA, and with the University of Pennsylvania, Philadelphia, PA, USA.
He has authored or coauthored >200 publications in conferences and impacted journals, including IJRR, AURO, JFR, ASC, EJC, with >7700 citations indexed by Scholar and h-index 51.
His team won multiple robotic challenges in MBZIRC 2017, MBZIRC 2020, and DARPA SubT competitions.
\end{IEEEbiography}

\EOD

\end{document}